\documentclass{article} 
\usepackage{iclr2027_conference,times}

\usepackage{amsmath,amsfonts,bm}

\def\eqref#1{equation~\ref{#1}}

\def\1{\bm{1}}

\DeclareMathAlphabet{\mathsfit}{\encodingdefault}{\sfdefault}{m}{sl}
\SetMathAlphabet{\mathsfit}{bold}{\encodingdefault}{\sfdefault}{bx}{n}

\usepackage{hyperref}
\usepackage{url}
\usepackage{graphicx}
\usepackage{wrapfig}
\usepackage{array}          
\usepackage[table]{xcolor}  
\usepackage{multirow}       
\usepackage{booktabs}       

\usepackage{siunitx}
\usepackage{tabularx}

\title{Latent Lattice Language Models}

\author{%
Shivam Duggal\textsuperscript{1,2}\thanks{Work done during summer internship at Adobe.}
\quad Zongze Wu\textsuperscript{2}
\quad William T. Freeman\textsuperscript{1} \quad
Antonio Torralba\textsuperscript{1} \\
\textbf{Richard Zhang\textsuperscript{2}
\quad Eli Shechtman\textsuperscript{2}} \\
\textsuperscript{1}MIT CSAIL \qquad \textsuperscript{2}Adobe
}

\iclrfinalcopy 
\begin{document}

\maketitle

\begin{abstract}
Language models typically increase per-token computation
through deeper or looped layers and latent reasoning steps, while richer
residual connections improve information flow across layers.
We instead increase \textit{computational capacity}: the number of active
computational states a model maintains.
We introduce the \textbf{Lattice of Latents},
a set of per-layer \textit{horizontal latent streams}
that run recurrently along time, alongside the vertical residual stream
that flows through depth.
Like hyper-connections, mHC, and attention or value residuals,
the Lattice enhances the residual stream, but rather than
adding or rerouting residual pathways through depth,
it provides \textit{companion computational states} that persist across tokens.
Unlike hybrid Transformer--SSM architectures,
whose recurrent states primarily serve as internal
memory for sequence mixing, these latents directly condition the
vertical stream's attention, KV-cache writes, MLP computation,
and decoding.
On Transformers trained on 20B FineWeb-Edu tokens, a 406M-parameter
Lattice model lowers validation perplexity from $15.41$ to $14.40$ { over a parameter-matched Transformer}.
These gains extend across architectural variants.
The Lattice composes with prior residual-stream enhancements:
with hyper-connections and input residuals,
it lowers validation perplexity from $15.41$ (base Transformer) and
$14.22$ (Transformer with the same enhancements) to $13.61$.
On Gated DeltaNet, Mamba-2, Triadic GDN,
and GDN--Mamba-2 hybrids, it improves perplexity
at every tested native state size,
with Lattice-GDN and Lattice-Triadic-GDN
variants outperforming pure baselines with $4\times$ larger
native states.
These results identify persistent per-layer state as a
complementary axis for scaling language-model computation.
{\textbf{Project Page:}
\hypersetup{urlcolor=magenta}
\textcolor{magenta}{\url{https://shivamduggal4.github.io/latent-lattice-language-models/}}}
\end{abstract}

\section{Introduction}

Language models accept raw text as input, transform it through internal representations,
and predict subsequent text. Like most learned systems, this process relies on some form
of latent computation. But what, precisely, is the latent computation in a large language
model?

In a standard autoregressive LLM, each token is processed by a stack of
layers along a single residual stream, and this one latent does several jobs at once.
At every layer, it reads from memory through attention queries, writes to memory
through keys and values, updates itself through the MLP, and, at the top, is decoded
into the next token.
The memory it writes to, the KV cache, is an append-only record that stays fixed once
written; the residual stream carries all of the active computation.
Recent work increases the computation performed on this stream:
deeper or looped layers apply more updates per
token~\cite{dehghani2019universaltransformers, geiping2025scalingtesttimecomputelatent, ouro},
continuous latent reasoning continues latent processing between
tokens~\cite{coconut, pccot, ponderlm2, CODI},
and richer residual connections, such as hyper-connections, mHC, and attention or
value residuals, improve how the stream carries information across
depth~\cite{zhu2025hyperconnections, xie2026mhcmanifoldconstrainedhyperconnections, kimiteam2026attentionresiduals, zhou2025valueresiduallearning}.

\begin{wrapfigure}{r}{0.6\columnwidth}
\centering
\vspace{-1mm}
\includegraphics[width=\linewidth]{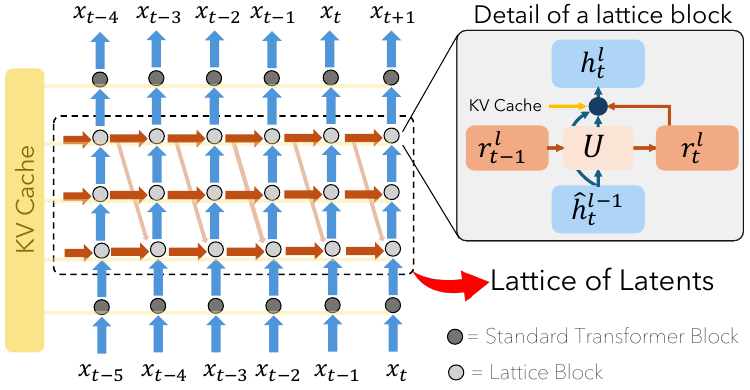}
\caption{\textbf{Latent Lattice Language Model.}
Blue arrows denote the vertical latent $h_t^\ell$ flowing through depth;
orange arrows denote the horizontal latents $r_t^\ell$, one per Lattice layer,
flowing through time.
The arrow from the top Lattice layer to the bottom layer of the next timestep illustrates the optional top-to-bottom feedback; the other two enhancements (input injection and the depth-wise readout stream) are omitted for clarity.
See the project webpage for animations and a comparison with the base Transformer.}
\label{fig:teaser_flow}
\vspace{-3mm}
\end{wrapfigure}

These approaches give the residual stream more to compute.
We focus instead on increasing \textit{computational capacity}: the number of active
computational states a model maintains.
We propose a \textit{lattice of computational latents}: alongside the vertical latent
$h_t^\ell$ that flows through depth (blue arrows in Fig.~\ref{fig:teaser_flow}),
each Lattice layer maintains its own horizontal latent $r_t^\ell$ (orange arrows),
updated recurrently across timesteps.
We call this architecture the \textbf{Lattice of Latents}, and a language model
built on it the \textbf{Latent Lattice Language Model (LLLM)}.
Like residual-stream enhancements~\citep{zhu2025hyperconnections, xie2026mhcmanifoldconstrainedhyperconnections, kimiteam2026attentionresiduals}
, the Lattice augments the vertical
stream, but rather than widening or rerouting it through depth, it adds companion
computational states that persist across tokens.
The horizontal latents add fixed-size recurrent state without 
additional
token positions or 
KV-cache entries. So, how does this horizontal flow of information support the vertical latent?

At each step of computation -- layer $\ell$, timestep $t$ -- two streams
of active information are available: the vertical latent $h_t^{\ell-1}$ from the
layer below, and the horizontal latent $r_t^\ell$, formed by updating the same
layer's previous state $r_{t-1}^\ell$.
In our Transformer instantiation, both latents jointly drive the layer's attention
and MLP computations. The horizontal latent conditions attention queries and
KV-cache writes, modulates the MLP, and, at the topmost Lattice layer, also
contributes to output decoding alongside the final vertical latent.
We design the horizontal update to be
(a) expressive enough to serve as a useful companion to the residual stream,
and (b) compatible with token-parallel, teacher-forced training.

Towards these goals, we start with an affine
Mamba-style~\citep{gu2024mambalineartimesequencemodeling} update in which
$r_t^\ell$ is updated from $r_{t-1}^\ell$ using the incoming vertical latent $h_t^{\ell-1}$.
Conditional on the layer-input sequence, this affine recurrence admits an exact
associative scan across token positions.
Although the update resembles those in Mamba and hybrid
Transformer--SSM models, \textit{its role differs}.
There, the recurrent state acts as a sequence mixer: an internal memory that replaces
attention or operates alongside it.
Here, attention (or the backbone's native memory) remains the sequence mixer, and the
horizontal latent acts as a \textit{companion to the residual stream}, conditioning
how the vertical stream reads and writes that memory and how its MLP processes it.

In this base update, $r_{t-1}^\ell \to r_t^\ell$, the horizontal latent
receives new information only through the incoming vertical latent (the residual stream) $h_t^{\ell-1}$.
We therefore add three enhancements that enrich the horizontal pathway:
\textit{input injection} feeds the token embedding directly into every layer's update,
so each horizontal latent sees the current token without relying on the vertical
stream to preserve it;
a nonlinear \textit{depth-wise readout stream} lets each layer's readout of its
horizontal latent depend on those of the layers below, adding nonlinear
cross-layer computation on top of the per-layer affine recurrences;
and \textit{top-to-bottom feedback} feeds the top layer's horizontal latent into the
bottom layer's update at the next timestep, so the most processed state
of the previous token informs the earliest computation on the current one.
The first two preserve the parallel scan, keeping our main models single-pass
at approximately matched compute. The top-to-bottom feedback does not preserve token parallelism, so we approximate it with a
few token-parallel Jacobi passes~\citep{pccot, ponderlm2}, at about $2.6\times$
the FLOPs for three passes during training; at inference, the same model can instead
be rolled out sequentially at the base FLOPs.

Across pretraining budgets from 10B to 30B FineWeb-Edu tokens, the Lattice of
Latents improves language-modeling quality at approximately matched parameters and
compute
(Tab.~\ref{tab:20b_arch_fineweb_ood}, Tab.~\ref{tab:30b_arch_fineweb_ood},
Fig.~\ref{fig:dl_ablation_2}).
At 20B tokens, a 406M-parameter LLLM with input injection and the depth-wise
readout stream lowers FineWeb-Edu perplexity from $15.41$ to $14.40$,
\emph{matching} a $1.5\times$ larger, 40-layer Transformer.
{A separately trained LLLM with the nonlinear top-to-bottom refiner, run with three Jacobi sweeps
(at higher FLOPs), lowers perplexity further to $14.03$ (Tab.~\ref{tab:20b_arch_fineweb_ood}).}
These gains extend to out-of-distribution perplexity, downstream tasks,
and finetuned mathematical reasoning (GSM8K).
As a memory-conditioning residual-stream enhancement, the Lattice alone
lowers perplexity more than hyper-connections, mHC, or value residuals, and it
composes with them (Sec.~\ref{sec:residual_experiments}): combined with
hyper-connections and input residuals, it reaches $13.61$, versus $14.22$ without
the Lattice (Fig.~\ref{fig:residual_stream_comparison}).

\definecolor{LLLMBlockGray}{RGB}{245,245,245}

\begin{table*}[t]
  \centering

  \small
  \setlength{\tabcolsep}{4.0pt}
  \renewcommand{\arraystretch}{1.12}
  \sisetup{
    table-number-alignment = center,
    group-digits           = false
  }

  \resizebox{\textwidth}{!}{%
  \begin{tabular}{@{}
    l
    S[table-format=3.1]
    S[table-format=1.2]
    S[table-format=1.2]
    S[table-format=2.3]
    S[table-format=2.3]
    S[table-format=2.3]
    S[table-format=2.3]
    S[table-format=2.3]
  @{}}
    \toprule

    \multirow{2}{*}{\textbf{Approach}}
      & \multicolumn{1}{c}{
          \multirow{2}{*}{
            \shortstack{
              \textbf{Params}\\[-0.2ex]
              \textbf{(M)}
            }
          }
        }
      & \multicolumn{2}{c}{\textbf{GFLOP/token}}
      & \multicolumn{1}{c}{
          \multirow{2}{*}{
            \shortstack{
              \textbf{FineWeb}\\\textbf{Edu}
            }
          }
        }
      & \multicolumn{1}{c}{
          \multirow{2}{*}{
            \shortstack{
              \textbf{Pile}
            }
          }
        }
      & \multicolumn{2}{c}{\textbf{LAMBADA}}
      & \multicolumn{1}{c}{
          \multirow{2}{*}{
            \shortstack{
              \textbf{Nemo}\\\textbf{CC}
            }
          }
        } \\

    \cmidrule(lr){3-4}
    \cmidrule(lr){7-8}

      & &
      {\textbf{Train}}
      & {\textbf{Infer}}
      &
      &
      &
      {\textbf{OAI}}
      & {\textbf{Stand.}}
      & \\

    \midrule

    \rowcolor{LLLMBlockGray}
    \multicolumn{9}{@{}l@{}}{%
      \rule{0pt}{2.7ex}%
      \textbf{400M-parameter models}%
    } \\

    Base-24L (width $d_{\text{model}}$$=1024$)
      & 406.7
      & 0.71
      & 0.71
      & 15.410
      & 26.276
      & 41.172
      & 73.648
      & 14.768 \\

    LLLM $(945,256)$
      & 406.5
      & 0.72
      & 0.72
      & 14.545
      & 25.031
      & 38.835
      & 65.780
      & 14.278 \\

    LLLM $(908,256)$ inj+vs
      & 405.6
      & 0.72
      & 0.72
      & 14.400
      & \textbf{24.100}
      & 38.513
      & 63.259
      & 14.116 \\

    LLLM-410 J3 $(945,256)$
      & 406.9
      & 1.85
      & 1.85
      & \textbf{14.026}
      & 24.107
      & \textbf{37.531}
      & \textbf{62.923}
      & \textbf{13.694} \\

    \addlinespace[1pt]

    \hspace{1.15em}
    \shortstack[l]{%
      \(\hookrightarrow\)\,\textit{sequential rollout at inference}\\[-0.25ex]
      \hspace{1.55em}\textit{(no Jacobi sweeps)}%
    }
      & \multicolumn{1}{c}{406.9}
      & \multicolumn{1}{c}{1.85}
      & 0.72
      & 14.733
      & 25.461
      & 38.737
      & 68.398
      & {\textemdash} \\

    \midrule

    \rowcolor{LLLMBlockGray}
    \multicolumn{9}{@{}l@{}}{%
      \rule{0pt}{2.7ex}%
      \textbf{500M-parameter models}%
    } \\

    LLLM $(512,512)$, 40L
      & 495.2
      & 0.94
      & 0.94
      & 14.368
      & 24.972
      & 38.119
      & 64.649
      & 14.127 \\

    \midrule

    \rowcolor{LLLMBlockGray}
    \multicolumn{9}{@{}l@{}}{%
      \rule{0pt}{2.7ex}%
      \textbf{600M-parameter models}%
    } \\

    Base-24L (width $d_{\text{model}}$$=1620$)
      & 603.3
      & 1.04
      & 1.04
      & 14.690
      & 24.443
      & 38.666
      & 68.939
      & 14.018 \\

    Base-24L (width $d_{\text{model}}$$=1472$)
      & 606.2
      & 1.06
      & 1.06
      & 14.649
      & 24.610
      & 39.272
      & 67.624
      & 14.221 \\

    Base-40L (width $d_{\text{model}}$$=1024$)
      & 609.1
      & 1.11
      & 1.11
      & 14.400
      & 24.234
      & 38.624
      & 67.508
      & 13.883 \\

    LLLM $(1024,1024)$
      & 603.0
      & 1.10
      & 1.10
      & \textbf{13.704}
      & \textbf{24.200}
      & \textbf{37.012}
      & \textbf{60.132}
      & \textbf{13.443} \\

    \bottomrule
  \end{tabular}%
  }

  \caption{
  \textbf{Language-modeling Perplexity ($\downarrow$) after 20B-token pretraining.} LLLM $(d_{\text{model}}, d_r)$ denotes a vertical (residual-stream) width $d_{\text{model}}$ and horizontal-latent width $d_r$; all models are 24 layers unless noted and use a 2048-token context. FineWeb-Edu is in-distribution; the other sets are out-of-distribution. The 400M LLLM $(908, 256)$ \texttt{inj+vs} (input injection + depth-wise readout stream) matches or outperforms the 600M Transformers. Results at 30B tokens are in Tab.~\ref{tab:30b_arch_fineweb_ood}.}
  \label{tab:20b_arch_fineweb_ood}
  \vspace{-5mm}
\end{table*}

The Lattice is not specific to softmax attention.
We also instantiate it on Gated DeltaNet, Mamba-2, Triadic GDN, and GDN--Mamba-2
hybrids, where the horizontal latents condition the backbones' native memory-read,
write, and update operations.
After 10B-token pretraining, at approximately matched parameter counts, the Lattice
improves perplexity at every native state size we test;
Lattice GDN and Lattice Triadic GDN each outperform pure baselines holding
$4\times$ as many native-state entries.
Downstream gains also hold across GDN, Mamba-2, and GDN--Mamba-2 hybrids:
adding the Lattice improves the nine-task average in nearly every
comparison, spanning five native-state budgets and both zero- and five-shot
settings.
Probing GDN's recurrent memory reveals a structural correlate of these gains:
at matched native-state sizes, Lattice models exhibit more concentrated state
spectra and greater query alignment with the leading memory direction.
These observations are consistent with more effective use of native memory.


Overall, comparisons at similar parameter and compute budgets show that allocating capacity
to both vertical and horizontal latents outperforms spending it on residual-stream
width or network depth alone in the tested configurations.
Together, these results identify persistent per-layer state as a
complementary axis for scaling language-model computation.

\vspace{-2mm}
\section{Related Work}
\vspace{-2mm}

\paragraph{More computation on a single stream.}
Recent work increases how much computation a language model applies to each token.
Along \emph{depth}, models add layers or repeatedly apply a shared stack
\citep{dehghani2019universaltransformers, geiping2025scalingtesttimecomputelatent, ouro};
along \emph{time}, continuous latent reasoning feeds the final hidden state back as the
next input, running additional latent steps before emitting a token
\citep{coconut, CODI, ponderlm2}, with Jacobi-style refinement recovering token-parallel
training \citep{ponderlm2, pccot}.
Both directions apply more updates to one active stream, and often grow persistent
memory with it: extra layers and loop-specific caches, or latent thoughts held as extra
positions, add KV-cache entries.
The Lattice instead increases computational capacity by maintaining one active latent per
Lattice layer that persists across tokens, without adding per-token KV-cache entries.

\vspace{-2mm}
\paragraph{Residual-stream enhancements.}
A complementary line of work changes how the residual stream carries information
through depth.
Hyper-connections expand it into parallel streams with learned mixing
\citep{zhu2025hyperconnections}, which mHC constrains to be doubly stochastic
\citep{xie2026mhcmanifoldconstrainedhyperconnections}; attention residuals replace the
fixed residual sum with attention over earlier layers' outputs
\citep{kimiteam2026attentionresiduals}; value residuals reuse first-layer values in
later layers \citep{zhou2025valueresiduallearning}; and input residuals re-inject the
token embedding before every layer.
These methods widen or reroute per-token computation through depth, but none gives a
layer its own state that persists from one token to the next.
The Lattice adds such states and composes with these enhancements: in our experiments,
adding it lowers perplexity on top of standard, input-residual, and hyper-connected
residual streams (Sec.~\ref{sec:residual_compose}).

\vspace{-2mm}
\paragraph{Recurrent sequence models and hybrids.}
Our affine horizontal update shares the scan-parallel form of linear recurrent
models such as Mamba \citep{gu2024mambalineartimesequencemodeling,
dao2024transformersssmsgeneralizedmodels}, Gated DeltaNet
\citep{yang2025gateddeltanetworksimproving}, and DeltaProduct \citep{deltaproduct},
whose transitions remain linear or affine in the previous state.
In these models, and in hybrids that replace attention with SSM layers at
selected depths \citep{lieber2024jambahybridtransformermambalanguage} or run them
alongside attention heads
\citep{dong2024hymbahybridheadarchitecturesmall}, the recurrent state is a sequence
mixer: it substitutes for or supplements attention as the means of reading the past.
In the Lattice, the horizontal latent is instead a companion to the residual stream:
in Transformers, softmax attention is retained, and the latent conditions attention, its KV-cache writes,
and the MLP.
Because this role is distinct from sequence mixing, the Lattice also complements
recurrent backbones: on GDN, Mamba-2, Triadic GDN, and GDN--Mamba-2 hybrids, it lowers
perplexity at approximately matched parameter counts (Sec.~\ref{sec:recurrent_backbones}).

\vspace{-2mm}
\paragraph{Recurrence within Transformers.}
Other work brings recurrence into Transformers without replacing attention.
{\citet{mozer2026topologicaltroubletransformers} argue that feedforward Transformers push evolving state into ever-deeper layers, exhausting depth, and call for recurrence.}
NextLat \citep{teoh2026nextlatentpredictiontransformerslearn} keeps the architecture unchanged and adds an auxiliary
objective that trains a latent dynamics model to predict the next hidden state,
encouraging recurrent-like belief states; the Lattice instead adds explicit recurrent
states to the architecture.
Several concurrent works add recurrent or multi-stream pathways, typically feeding a
deeper representation from the previous token back into the current token's computation
\citep{wang2026fullbandwidthtransformer, huang2026latentrecurrenttransformerarchitecture,
cai2026t2mlrtransformertemporalmiddlelayer, mozer2026recirculation}, or separating
state preparation from next-token prediction into distinct streams
\citep{monea2026statepredictionseparationhypothesis}.
Among these, T$^2$MLR \citep{cai2026t2mlrtransformertemporalmiddlelayer} is closest to
our Jacobi refiner, though its recurrent object is a cached vertical representation
rather than a lattice of layer-wise horizontal states.
These works largely add a single feedback pathway or separate two roles, whereas the
Lattice maintains a persistent latent at every Lattice layer;
Appendix~\ref{app:related_work} gives detailed description of concurrent works.

\section{Method}
\label{sec:method}

\begin{figure}[t]
    \centering
    \includegraphics[width=\linewidth]{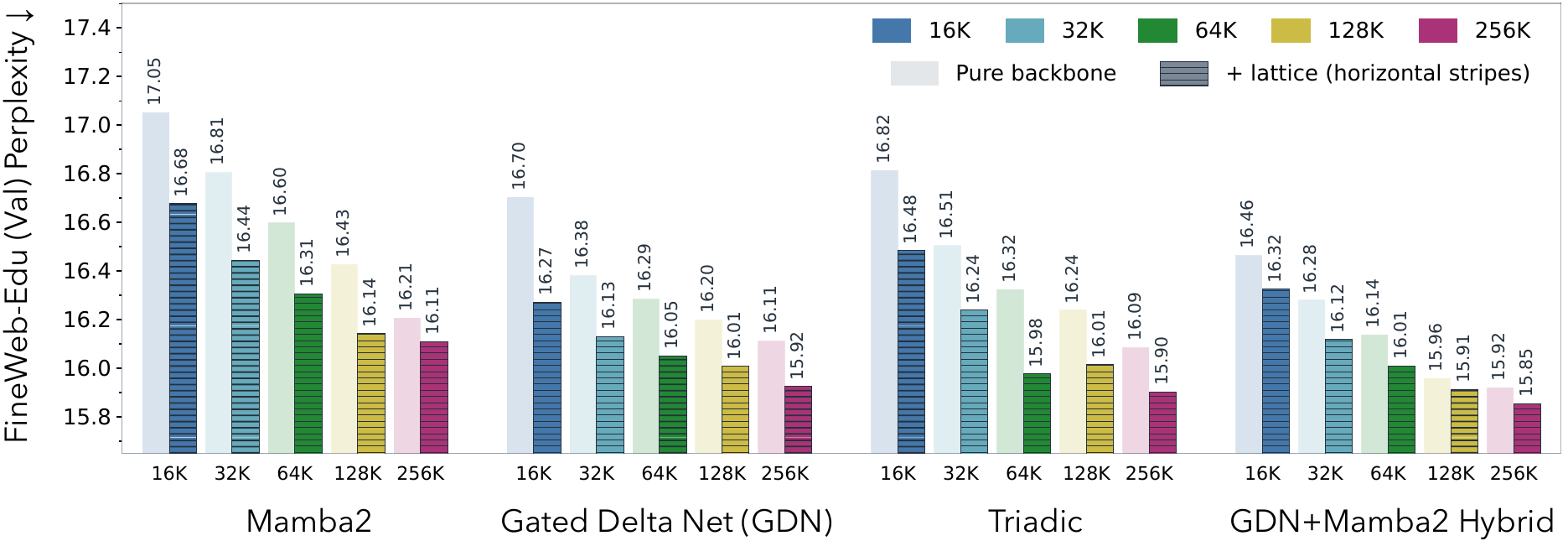}
    \vspace{-4mm}
    \caption{\textbf{Recurrent backbones: FineWeb-Edu perplexity.}
       Pure and Lattice GDN, Mamba2, Triadic GDN, and GDN--Mamba2 hybrid models ($\sim$405M parameters, 10B FineWeb-Edu tokens) across multiple native-state sizes per mixer. Each Lattice model is iso-parameter with its pure counterpart and lowers perplexity at every state size.
       See Sec.~\ref{sec:recurrent_backbone_experiments}.}
    \label{fig:recurrent_backbone_fineweb_results}
    \vspace{-2mm}
\end{figure}

\vspace{-2mm}
We propose the \textbf{Lattice of Latents}, an augmentation that distributes
computation across a lattice of latents rather than relying only on the usual
single residual-stream latent per token. The latent computation spans two axes.
Along the \emph{vertical} axis (network depth), the per-token latent is initialized
from the current token embedding and flows bottom-to-top through the backbone's
layers, where it is updated by attention or recurrent state computations followed
by a position-wise MLP, before being decoded into the next-token distribution.
Along the \emph{horizontal} axis (time), the Lattice of Latents adds one persistent
latent vector at each layer of an intermediate Lattice block, evolving across
timesteps rather than through depth. Each horizontal latent remains at its own
layer and abstraction level, assisting the vertical stream by conditioning that
layer's memory reads and updates. In our Transformer instantiation, we additionally
study MLP modulation.
The Lattice thereby increases \textit{computational capacity} –– the
number of active computational states a model maintains by adding companion
states to the residual stream, which condition the computation of the residual stream, rather than simply widening the residual stream or adding residual pathways through depth.

We integrate the Lattice of Latents into standard Transformers,
Transformers with residual-stream enhancements (hyper-connections, mHC, value residuals, and input
residuals), and recurrent backbones: Gated DeltaNet (GDN), Mamba-2, Triadic GDN,
and GDN--Mamba-2 hybrids.
We begin with background on the per-token computational graph of the vertical
latent (Sec.~\ref{sec:background}), then formulate the Lattice of Latents
(Sec.~\ref{sec:lattice}), its composition with residual-stream enhancements
(Sec.~\ref{sec:residual_compose}), and its extension to recurrent backbones
(Sec.~\ref{sec:recurrent_backbones}).

\begin{figure}[t]
    \centering
    \includegraphics[width=\linewidth]{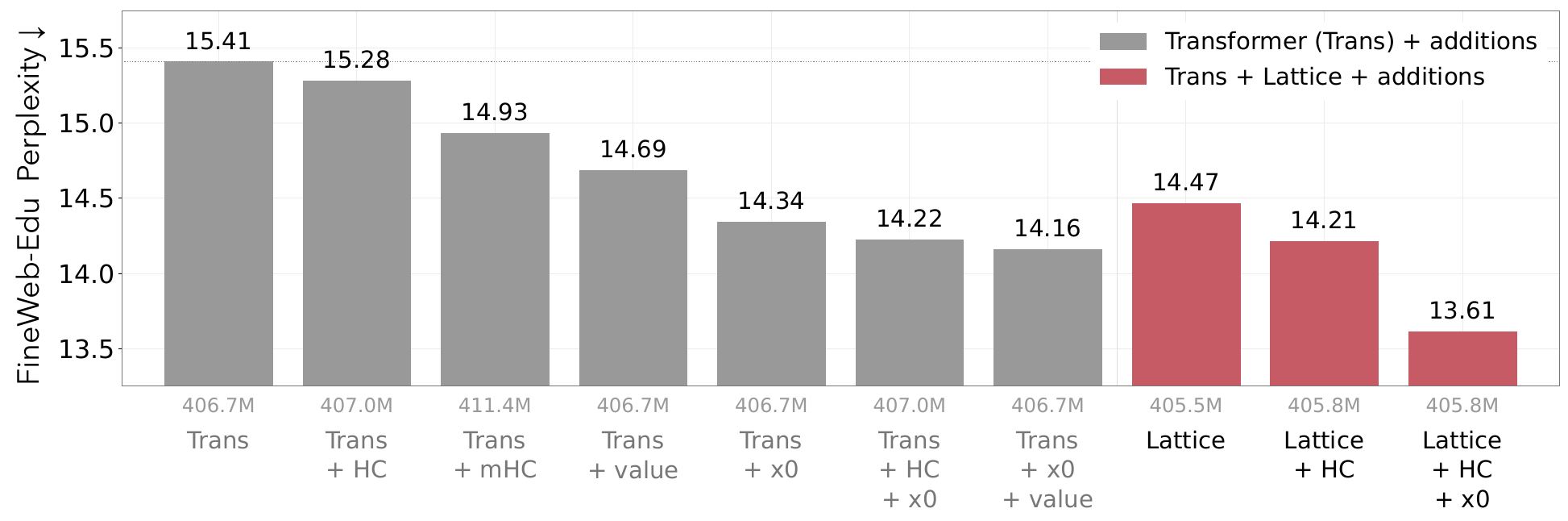}
    \vspace{-7mm}
    \caption{\textbf{{Lattice of Latents vs.\ prior residual-stream enhancements.}} {On its own, the} Lattice {lowers the base Transformer's} perplexity more than {HC, mHC, or} value residuals, and {its gains compound with theirs:} Lattice\,+\,HC\,+\,x0 {reaches} $13.76${, the best overall. 400M models, 20B FineWeb-Edu tokens. See Sec.~\ref{sec:residual_experiments} and Appendix Sec.~\ref{app:residual_streams}.}}
    \label{fig:residual_stream_comparison}
    \vspace{-2mm}
\end{figure}

\subsection{Background: Computation with a Single Active Latent}
\label{sec:background}

\paragraph{The per-token computational graph of the vertical latent.}
Consider an autoregressive Transformer of depth $N$ and width $d_{\mathrm{model}}$, processing
a token stream $x_1,x_2,\dots$. At timestep $t$, the model maps the input token
$x_t$ to a distribution over the next token $x_{t+1}$ through a single
residual-stream latent $h_t^\ell\in\mathbb{R}^{d_{\mathrm{model}}}$, initialized from the token
embedding $h_t^0$ and refined across the $N$ layers.
Layer $\ell\in\{1,\dots,N\}$ updates this
latent with a pre-norm attention block followed by an MLP block:
\vspace{-1mm}
\begin{align*}
q_t^\ell,\,k_t^\ell,\,v_t^\ell
&=W_q^\ell\hat h,\;W_k^\ell\hat h,\;W_v^\ell\hat h\\
\bar h_t^\ell
&=h_t^{\ell-1}+W_o^\ell\,\mathrm{Attn}\!\big(q_t^\ell,\mathcal{M}_t^\ell\big)\\
h_t^\ell
&=\bar h_t^\ell+\mathrm{MLP}^\ell\!\big(\mathrm{Norm}(\bar h_t^\ell)\big),
\end{align*}
where $\hat h=\mathrm{Norm}(h_t^{\ell-1})$, $\mathrm{Norm}$ is RMSNorm and
$\mathcal{M}_t^\ell=\{(k_s^\ell,v_s^\ell)\}_{s\le t}$ is the layer-$\ell$ KV cache
to which each token appends its keys and values.
The final latent is decoded as
$p(x_{t+1}\mid x_{\le t})=\mathrm{softmax}\big(W_U\,\mathrm{Norm}(h_t^N)\big)$.
Thus one latent per token does all the active computation (reading memory, writing to it, and producing the output), while the KV cache passively stores its past projections.

\vspace{-2mm}
\paragraph{The recurrent view of the vertical latent.}
Continuous latent reasoning~\citep{coconut,CODI,ponderlm2} extends this
pathway across timesteps by feeding the previous step's final-layer latent back as
the next input, $h_t^0=h_{t-1}^N$,
bypassing decoding and re-embedding.
The feedback can thus carry more than the $\log_2 V$ bits of a discrete token,
and the vertical latent becomes a running state across time.
A standard Transformer instead re-initializes its vertical latent from the token
embedding at every timestep.


\subsection{The Lattice of Latents}
\label{sec:lattice}

How can we increase the computational capacity of this single active latent?
One could widen it (larger $d_{\mathrm{model}}$), replicate it into parallel depth-wise streams as in
hyper-connections, or add latents along a different axis.
We take the third route: each Lattice layer $\ell$ maintains its own
\emph{horizontal} latent $r_t^\ell$ that runs across timesteps, initialized from a
learned state and updated from its previous value $r_{t-1}^\ell$ and the incoming
vertical latent $h_t^{\ell-1}$.
The vertical latent advances through depth at each timestep and the horizontal
latents advance through time at each layer; together they form a lattice.
We compare this against widening and show that it composes with residual-stream enhancements.
Below we describe how the horizontal latents act
(Sec.~\ref{sec:horizontal_act}) and how they are updated
(Sec.~\ref{sec:horizontal_update_sec}).

\vspace{-2mm}
\subsubsection{How the horizontal latents act}
\label{sec:horizontal_act}

\begin{wrapfigure}{r}{0.3\columnwidth}
    \centering
    \vspace{-8mm}
    \includegraphics[width=\linewidth]{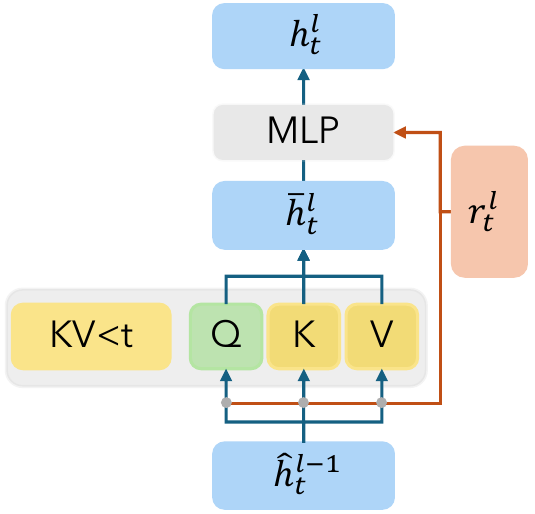}
    \caption{\textbf{How the horizontal latent acts.}
    Transformer instantiation: the recurrent lattice latent $\hat r_t^\ell$
    conditions the layer's QKV and MLP computations; when the optional
    horizontal decoding head is enabled, the topmost horizontal latent also
    contributes to output decoding (not shown).}
    \label{fig:horizontal_act}
    \vspace{-8mm}
\end{wrapfigure}

\vspace{-1mm}
Let $r_t^\ell\in\mathbb{R}^{d_r}$ be the horizontal latent at layer $\ell$ and
timestep $t$, with width $d_r$ chosen independently of $d_{\mathrm{model}}$.
A standard Transformer layer computes from the vertical latent $h_t^{\ell-1}$ and its
KV cache. In the Lattice, layer $\ell$ has \emph{two} latent inputs: the vertical
latent $h_t^{\ell-1}$ from the layer below, and the horizontal latent $r_t^\ell$,
updated from the same layer's previous state $r_{t-1}^\ell$.
Together they condition the layer's attention and MLP, and thus both the residual
update that produces $h_t^\ell$ and the keys and values written to memory.
Optionally, the topmost Lattice layer's horizontal latent also contributes to
decoding. We detail each contribution below (Fig.~\ref{fig:horizontal_act});
all our Lattice Transformers modulate both attention and the MLP.
All contributions read the normalized latent $\hat r_t^\ell=\mathrm{Norm}(r_t^\ell)$.

\textbf{Conditioning Attention via FiLM.}
The horizontal latent $\hat r_t^\ell$ modulates the query, key, and value projections of
$\hat h=\mathrm{Norm}(h_t^{\ell-1})$ via {Feature-wise Linear Modulation (FiLM)}~\citep{perez2017filmvisualreasoninggeneral}:
\begin{align*}
q_t^\ell
&= \big(W_q^\ell\hat h\big)\odot\big(1+\gamma_q(\hat r_t^\ell)\big)
   +\beta_q(\hat r_t^\ell)\\
\gamma_{(\cdot)}(\hat r_t^\ell)
&=\tanh\big(W^\gamma_{(\cdot)}\hat r_t^\ell\big)
\qquad
\beta_{(\cdot)}(\hat r_t^\ell)=W^\beta_{(\cdot)}\hat r_t^\ell
\end{align*}

and likewise for $k_t^\ell$ and $v_t^\ell$.
The multiplicative gain $1+\gamma\in(0,2)$ is bounded, while $\beta$ is an additive
shift; each of $q,k,v$ has its own layer-specific, zero-initialized $W^\gamma$ and
$W^\beta$, so the modulation starts as the identity.
The gain $\gamma$ may be shared across channel groups within each head, while $\beta$
is always per-channel.
Because the modulated $k_t^\ell,v_t^\ell$ are written to $\mathcal{M}^\ell$, the
horizontal latent shapes both the memory read and the memory write.
The attention output, projected by $W_o^\ell$, is added to the vertical latent as
before, giving the MLP input $\bar h_t^\ell$.

\textbf{Modulating MLP.}
The horizontal latent also shifts the gate and up pre-activations of the
layer's SwiGLU MLP, each through its own zero-initialized map:
\begin{align*}
m &= \mathrm{Norm}(\bar h_t^\ell)
\qquad
\mathrm{gate}=W_{\mathrm{gate}}^\ell m+\beta_{\mathrm{gate}}(\hat r_t^\ell)\\
\mathrm{up}
&=W_{\mathrm{up}}^\ell m+\beta_{\mathrm{up}}(\hat r_t^\ell).
\end{align*}
MLP output
$W_{\mathrm{down}}^\ell\big(\mathrm{SiLU}(\mathrm{gate})\odot\mathrm{up}\big)$
is added to $\bar h_t^\ell$, giving $h_t^\ell$.

\textbf{Modulating Decoding.}
Optionally, the topmost Lattice layer's normalized horizontal latent and the final
vertical latent are projected into a shared readout space, and the next token is
decoded from their sum (details in Appendix~\ref{app:decoding_modulation}).
We use this head only with a highly compressed vertical latent
($d_{\mathrm{model}}{=}256$) (in Fig.~\ref{fig:dl_ablation_3}), where the standard decoder would otherwise read the output distribution from a 256-dimensional latent alone.

\subsubsection{How the horizontal latents are updated}
\label{sec:horizontal_update_sec}

We divide the backbone's layers into an early block (E), an intermediate
\emph{Lattice} block (L), and a late block (D); horizontal latents exist only in L. For $\ell\in L$, the update takes the form:
\begin{equation}
r_t^\ell
=U\big(r_{t-1}^\ell;\;h_t^{\ell-1},\;h_t^0,\;r_{t-1}^{\mathrm{top}}\big)
\notag
\end{equation}

\begin{wrapfigure}{r}{0.45\columnwidth}
  \centering
  \vspace{-4mm}
  \includegraphics[width=\linewidth]{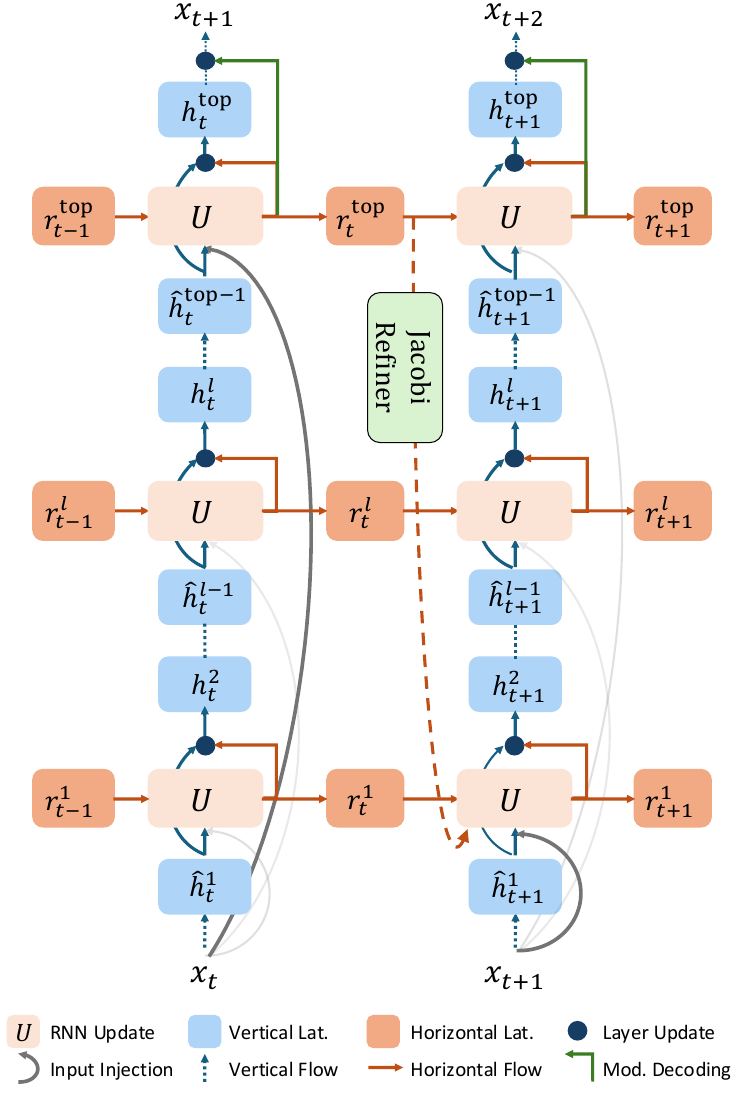}
  \vspace{-5mm}
  \caption{\textbf{Horizontal latent update.}
  On top of the Mamba-style affine update, we further introduce input injection,
  vstream (not shown) and an optional cross-timestep nonlinear flow.
  See webpage for animations.}
  \vspace{-27mm}
  \label{fig:horizontal_update}
\end{wrapfigure}

The base update uses the layer's own history $r_{t-1}^\ell$ and the incoming
vertical latent $h_t^{\ell-1}$.
Input injection adds the token embedding $h_t^0$, and the optional Jacobi refiner adds
feedback $r_{t-1}^{\mathrm{top}}$ from the top Lattice layer one timestep back,
injected only at the bottom Lattice layer and taken from the previous sweep.
The depth-wise readout stream (\texttt{vstream} or \texttt{vs} in model labels), which passes information upward from lower horizontal latents within a timestep, enriches each
layer's readout rather than entering $U$.
We describe each mechanism below (Fig.~\ref{fig:horizontal_update}).

\textbf{Affine update from $r_{t-1}^\ell$ to $r_t^\ell$.}
To keep training token-parallel and teacher-forced, we need a recurrence that admits
an exact associative scan across timesteps rather than a token-by-token loop.
An affine transition with input-dependent coefficients has this property.
Following linear attention and Mamba-style state-space models, we use one whose
coefficients depend only on the current layer input, never directly on the previous
horizontal state.
The affine update reads a normalized copy of the vertical latent $n_t^\ell=\mathrm{Norm}_{\mathrm{in}}(h_t^{\ell-1})$, and is defined as:

\begin{align*}
\Delta_t^\ell
&=\mathrm{softplus}\big(W_\Delta^\ell n_t^\ell+b_\Delta^\ell\big)\\
g_t^\ell
&=\exp\big(-\exp(a^\ell)\odot\Delta_t^\ell\big)\\
c_t^\ell
&=\mathrm{SiLU}\big(W_r^\ell n_t^\ell+b_r^\ell\big)\\
b_t^\ell
&=(1-g_t^\ell)\odot c_t^\ell\\
r_t^\ell
&=g_t^\ell\odot r_{t-1}^\ell+b_t^\ell.
\end{align*}

Here $r_0^\ell$ is a learned initial state before the first token $x_1$, and $a^\ell$
and $b_\Delta^\ell$ are learned per-channel log-decay rates and time-step biases.
Because $g_t^\ell$ and $b_t^\ell$ depend on $n_t^\ell$ alone, the transition is exactly
affine given the layer-input sequence, so the horizontal states at all token positions
are computed with a parallel scan during training.
Layers are still evaluated in depth order, each scanning the vertical inputs produced
by the layer below.
The readout used downstream is $\hat r_t^\ell=\mathrm{Norm}(r_t^\ell)$.
Although the update follows a Mamba-style gated affine recurrence, the horizontal
latent plays a different role, a companion to the residual stream rather than a
sequence mixer, which we revisit for backbones that already maintain recurrent memory
(Sec.~\ref{sec:recurrent_backbones}).

We call this configuration the \textbf{base update} and label the corresponding models
$\mathbf{LLLM}(d_{\mathrm{model}},d_r)$, with vertical width $d_{\mathrm{model}}$ and horizontal width $d_r$; additional
mechanisms are marked separately in model labels.
The base update can \emph{integrate} history across time, but on its own it applies no
learned nonlinear transformation to its previous state.
We therefore study three complementary mechanisms: input injection (\texttt{inj}) and
the depth-wise readout stream (\texttt{vs}) add expressivity within a timestep while
preserving the parallel scan, whereas the Jacobi refiner adds a nonlinear
top-to-bottom feedback pathway across timesteps at the cost of the single-pass
guarantee.

\textbf{Adding the input embedding to the horizontal latent.}
The base update sees the current input only through $h_t^{\ell-1}$, which by layer
$\ell$ may have discarded information about the raw token identity
(see the narrow-vertical-latent ablation, Fig.~\ref{fig:dl_ablation_3}).
\textbf{Input injection} (\texttt{inj}) also feeds the normalized token embedding $h_t^0$ into $\Delta_t^\ell$ and $c_t^\ell$ through zero-initialized projections. These coefficients still depend only on current-timestep inputs, so the transition remains exactly affine and the same parallel scan applies (more details in Appendix~\ref{app:input_injection}).

\textbf{Enhancing layer-wise nonlinearity of the horizontal latent.}
At layer $\ell$, the readout $\hat r_t^\ell$ from the base update and input
injection receives information from lower Lattice layers' horizontal latents only
indirectly, through whatever survives in the vertical latent.
The \textbf{depth-wise readout stream} (\texttt{vstream} or \texttt{vs}) adds a direct path: a purely
depth-wise state $s_t^\ell\in\mathbb{R}^{d_r}$ that runs bottom to top across the
Lattice, seeded with $s_t^{\ell_0-1}:=0$ at the first Lattice layer $\ell_0$:
\begin{align*}
s_t^\ell
&=s_t^{\ell-1}
+W_{s\uparrow}^\ell\mathrm{SiLU}\Big(
W_{s\downarrow}^\ell\big[
\mathrm{Norm}(\hat r_t^\ell),\,
\mathrm{Norm}(s_t^{\ell-1})
\big]\Big),
\end{align*}
where $[\cdot,\cdot]$ denotes concatenation.
Since $W_{s\uparrow}^\ell$ is zero-initialized, $s_t^\ell\equiv s_t^{\ell-1}$ at
initialization, so with the zero seed the stream is zero throughout the Lattice.
The stream carries same-timestep communication across depth, separately from the
recurrent update $U$, which acts across time.
Because $s_t^\ell$ depends only on same-timestep quantities, it adds a nonlinear
dependency across depth but none across time, leaving the base recurrence and its
parallel scan untouched.
With \texttt{vs} enabled, the enriched readout
\begin{align*}
\tilde r_t^\ell:=\hat r_t^\ell+\mathrm{Norm}(s_t^\ell)
\end{align*}
replaces $\hat r_t^\ell$ wherever it conditions attention or the MLP
(Sec.~\ref{sec:horizontal_act}).
The state carried across timesteps remains the raw $r_t^\ell$, not $\tilde r_t^\ell$;
the enriched readout influences higher layers' updates only indirectly, through the
vertical stream.

\textbf{Cross-timestep nonlinear update.}
Input injection and the depth-wise readout stream use only current-timestep
information, so neither feeds a \emph{higher} layer's more deeply processed latent
back to a \emph{lower} layer at the next timestep.
In the Transformer instantiation, we add this top-to-bottom feedback: the top Lattice
layer's latent at $t-1$, passed through a zero-initialized SwiGLU, is added to the
bottom Lattice layer's update at $t$, making the recurrence nonlinear across
timesteps.
This feedback is straightforward in a token-by-token rollout but breaks single-pass
teacher-forced training, so we approximate it with a \emph{Jacobi refiner}
\citep{ponderlm2,pccot}: after an initial pass over the Lattice block, each
refinement sweep reruns all Lattice layers in parallel across tokens,
taking the feedback from the previous sweep.
Unlike \citet{ponderlm2,pccot}, which apply Jacobi iteration to the \emph{vertical}
latent ($h_t^0=h_{t-1}^N$), we apply it to the \emph{horizontal} latents.
\texttt{J0} denotes the single-pass model without this cross-timestep
feedback and \texttt{J3} the initial pass followed by two refinement sweeps
($\approx2.6\times$ the FLOPs).
At inference, the feedback can instead be computed exactly by sequential rollout;
we evaluate both (details in Appendix~\ref{app:jacobi}).

\subsection{Composing with Residual-Stream Enhancements}
\label{sec:residual_compose}
Because the Lattice conditions each block's computation rather than the residual
pathway itself, it combines with residual-stream enhancements without changing the
horizontal update.
An \textit{input residual} re-injects the normalized token embedding before every
layer, $h_t^{\ell-1}\leftarrow\lambda_r^\ell h_t^{\ell-1}+\lambda_0^\ell\,
\mathrm{Norm}(h_t^0)$, with learned per-layer scalars; the horizontal update then reads
this re-injected layer input as $h_t^{\ell-1}$.
This differs from input injection (\texttt{inj}), which feeds the embedding into the
horizontal update rather than into the residual stream.
Hyper-connections~\citep{zhu2025hyperconnections} expand the residual stream into
$n$ parallel streams that a learned router mixes into each block's input and remixes
after its output (mHC~\citep{xie2026mhcmanifoldconstrainedhyperconnections}
constrains this mixing to be doubly stochastic).
With hyper-connections, each Lattice layer's horizontal update reads the mixed block
input in place of $h_t^{\ell-1}$, FiLM conditions that block's QKV and MLP
computations as in Sec.~\ref{sec:horizontal_act}, and the router writes the block
outputs back to the streams.
A single horizontal latent per Lattice layer is shared across the $n$ streams.
A value residual~\citep{zhou2025valueresiduallearning} mixes each layer's attention values
with those of the first layer; in Lattice layers, the mix is applied after FiLM
conditioning and before attention.
In our experiments, we compare the Lattice against these residual-stream enhancements
and evaluate their combinations.

\vspace{-2mm}
\subsection{Lattice of Latents on Recurrent Backbones}
\label{sec:recurrent_backbones}

Although the horizontal update uses a Mamba-style recurrence, the Lattice also
complements backbones that already maintain recurrent memory.
We instantiate it on GDN, Mamba-2, Triadic GDN, and GDN--Mamba-2 hybrids.
Unlike a Transformer's growing KV cache, these backbones keep native memories of
fixed size: matrix-valued states per head in GDN and Mamba-2, and a
three-dimensional state per head in Triadic GDN, which adds a routing axis.
The Lattice leaves these memories intact and adds a separate horizontal latent that
conditions how they are used.\footnote{We use ``state'' and ``memory''
interchangeably for the backbones' native recurrent memories, which play a role
analogous to the KV cache. The horizontal latent is also recurrent state, but its
role is to condition the computations driven by the vertical residual stream.}

We keep the base update, with input injection and the depth-wise readout stream
as options, and apply FiLM to each backbone's memory read, write, and update
quantities: in GDN, Q, K, V, the retention ($\alpha$) and update-strength ($\beta$)
pre-activations, and the output gate; in Mamba-2, the read (C) and write (B)
projections, the input stream, the time-step pre-activation, and the output gate.
Triadic GDN additionally conditions its routing factors, and GDN--Mamba-2 hybrids follow the {settings of each layer's backbone}.
All modulation weights are zero-initialized, and modulation is applied before each
quantity's constraining transformation. These comparisons use unmodulated MLPs, the ordinary decoding head, and no Jacobi refiner; further details are in the Appendix.

To separate this from simply enlarging native memory, we compare models with and
without the Lattice across native-state budgets at approximately matched
parameters.
In GDN, for example, a 256-dimensional horizontal latent combined with 65{,}536
native state entries per layer achieves lower perplexity than a pure GDN with
262{,}144 entries.
We also analyze GDN's state geometry (spectra, effective and stable rank, and query
alignment with the leading memory direction) with and without the Lattice.
The distinction is functional: the horizontal latent is itself persistent state,
but rather than expanding the backbone's associative memory, it provides a learned,
history-dependent pathway that conditions the vertical stream's computation.
We hypothesize that this added computational capacity helps the model use its
native memory more effectively, complementing rather than replicating increases in
memory capacity; our experiments are consistent with this hypothesis.  
\section{Experiments}

\begin{figure}[t]
    \centering
    \includegraphics[width=\linewidth]{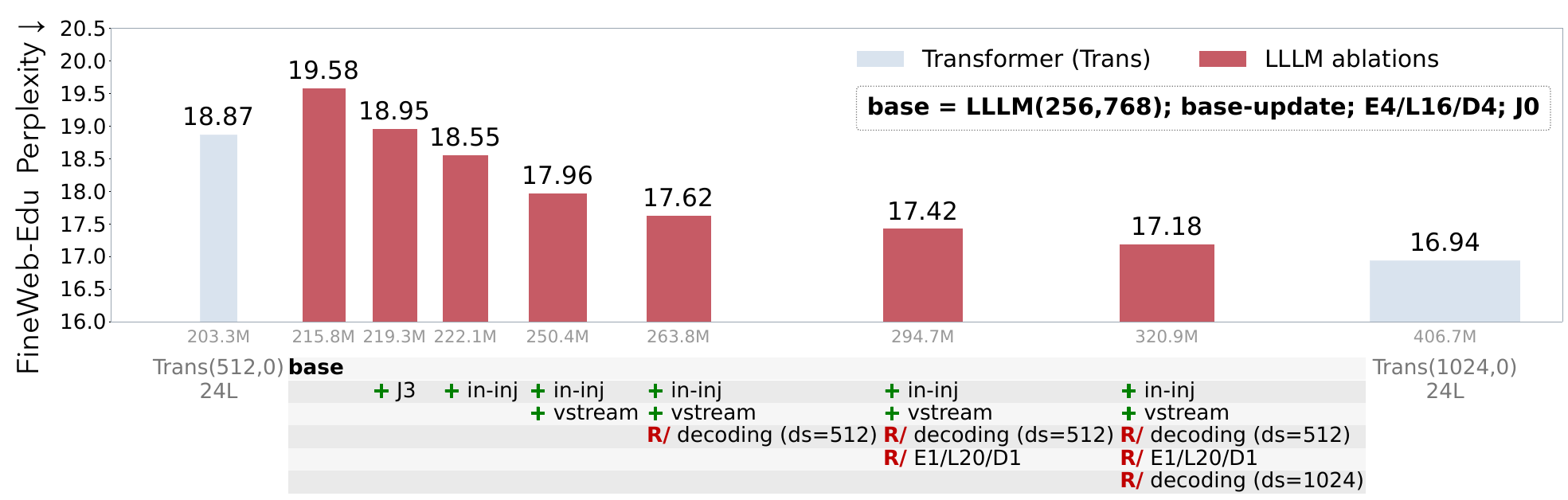}
    \vspace{-5mm}
    \caption{\textbf{Deconstructing the Lattice.} Here we study a challenging regime where the vertical residual stream width is significantly reduced to 256-dim. LLLM ablations show how the proposed horizontal update enhancements (Sec.~\ref{sec:horizontal_act} and Sec.~\ref{sec:horizontal_update_sec}) {recover most of} the lost performance{; trace} the red bars from left to right. Eval post 10B-token pretraining. Note: 406M Lattice model beats transformer (with perplexity  = 15.63, see Fig.~\ref{fig:dl_ablation_2}).
    See Sec.~\ref{sec:deconstructing_lattice} for more details.}
    \label{fig:dl_ablation_3}
    \vspace{-3mm}
\end{figure}

We evaluate the Lattice of Latents on language modeling and zero- and five-shot downstream tasks
for both Transformer and recurrent backbones, and combine it with residual-stream
enhancements.
We then deconstruct the Lattice into its components and study its effect on GDN's
native memory.
Further experiments on -- jacobi refiner, comparison with short conv, stateless modulation, downstream evaluations, GSM8K finetuning, and other details are in Appendix Sec.~\ref{sec:appendix_experiments}.

\subsection{Transformer Backbones}
\label{sec:transformer_backbones}

We pretrain 400M--600M-parameter LLLMs and Transformer baselines on FineWeb-Edu
for 10B, 20B, or 30B tokens; the larger budgets exceed
Chinchilla-style compute-optimal budgets at these sizes.
We report held-out perplexity on FineWeb-Edu (in-domain) and on Pile, LAMBADA
(OpenAI and Standard), and Nemotron-CC (out-of-domain), and zero- and
five-shot accuracy on nine lm-evaluation-harness tasks; evaluation and compute-accounting
details are in Appendix Sec.~\ref{sec:appendix_experiments}.

\paragraph{Language modeling.}
Table~\ref{tab:20b_arch_fineweb_ood} compares 400M--600M models pretrained on 20B
FineWeb-Edu tokens.
At 400M, LLLM$(945,256)$ matches Base-24L in parameters and forward compute
(406.5\,M / 0.72 vs.\ 406.7\,M / 0.71 GFLOP/token), yet lowers FineWeb-Edu perplexity from
$15.41$ to $14.54$ ($-5.6\%$) and out-of-domain perplexity by $3$--$11\%$ on every set.
Adding \texttt{input-injection}\,+\,\texttt{vstream}, with the vertical width reduced from
$945$ to $908$ to hold parameters and FLOPs approximately fixed, improves further, most
on out-of-domain data (the best Pile perplexity in the table, $24.10$).
Jacobi refinement gives the strongest 400M results: LLLM-410 J3 attains the best
FineWeb-Edu ($14.03$), LAMBADA, and Nemotron-CC perplexities in the 400M block, but its
three sweeps raise estimated forward compute to $1.85$ GFLOP/token.
Exact sequential rollout of the same checkpoint needs only $0.72$ GFLOP/token but raises
FineWeb-Edu perplexity to $14.73$.
Training with a randomized number of Jacobi iterations, as in
PonderLM-2~\citep{ponderlm2}, can remove this gap (Sec.~\ref{sec:deconstructing_lattice});
we train with three fixed sweeps.

At 600M, LLLM$(1024,1024)$ achieves the best perplexity on every dataset in its block
(FineWeb-Edu $13.70$), outperforming both width-scaled Transformers and the depth-scaled
Base-40L, which has nearly identical forward compute ($1.10$ GFLOP/token).
Thus, at comparable parameters and compute, splitting capacity between vertical and
horizontal latents outperforms spending it on residual-stream width or depth alone.
The advantage also holds across scales: the 405.6\,M \texttt{input-injection}\,+\,\texttt{vstream}
model matches the 609M Base-40L on FineWeb-Edu ($14.40$) and beats it on Pile and both
LAMBADA variants, trailing only on Nemotron-CC, with $33\%$ fewer parameters and $35\%$
less estimated forward compute.
Under three-sweep evaluation, the 407M J3 model surpasses Base-40L on all five
datasets, though at higher inference compute.
Results at 30B tokens are in Appendix Tab.~\ref{tab:30b_arch_fineweb_ood}.

\begin{figure}[t]
    \centering
    \includegraphics[width=\linewidth]{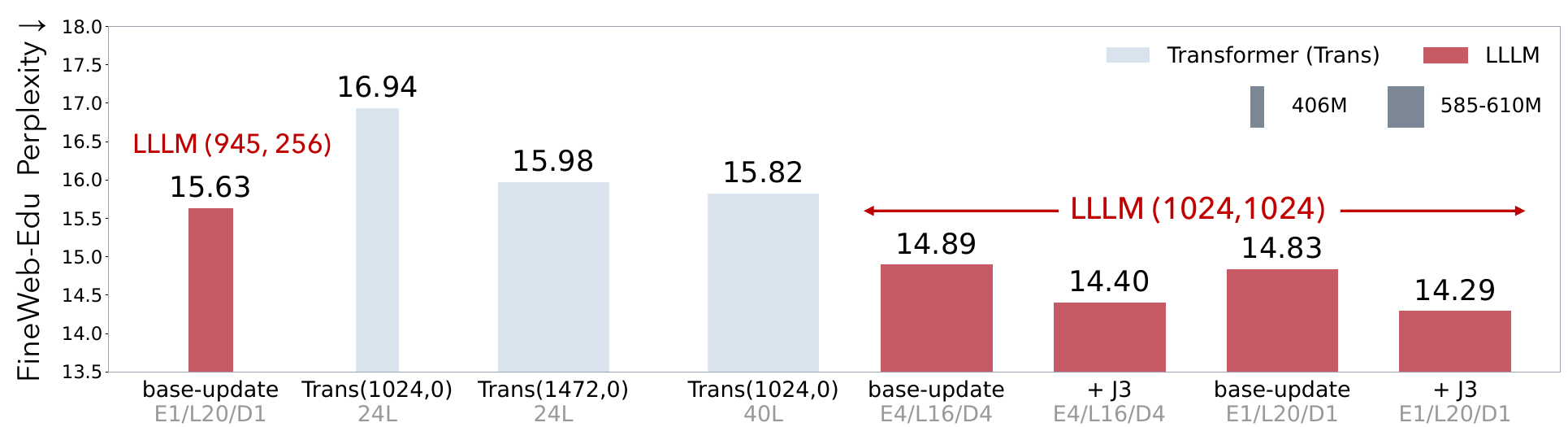}
    \vspace{-5mm}
    \caption{\textbf{LLLM vs.\ larger-depth and larger-width Transformers.} Acting as a companion to the residual stream, the Lattice enables better performance than a deeper Transformer or naively increasing the residual-stream width; even 406\,M LLLM$(945,256)$ outperforms both $\sim$600\,M alternatives on 10B-token pretraining.
    See Sec.~\ref{sec:deconstructing_lattice}.}
    \label{fig:dl_ablation_2}
    \vspace{-4mm}
\end{figure}

\vspace{-2mm}
\paragraph{Downstream evaluation.}
Table~\ref{tab:20b_downstream_zero_five_shot} reports the nine-task average accuracy
(Task Avg.) of the same 20B-token models.
At matched 400M scale, LLLM$(945,256)$ improves on Base-24L from $45.46$ to $47.67$
zero-shot ($+2.21$ points) and from $46.19$ to $47.79$ five-shot ($+1.60$).
Adding \texttt{input-injection}\,+\,\texttt{vstream} raises five-shot accuracy further to $48.16$,
and Jacobi refinement gives the strongest 400M averages, $48.69$ zero-shot and $49.38$
five-shot ($+3.23$ and $+3.19$ over Base-24L).
At 600M, LLLM$(1024,1024)$ outperforms Base-40L by $1.17$ points zero-shot and $2.00$
five-shot.
The {407M} J3 model also exceeds Base-40L, by $1.33$ and $2.01$ points, with roughly
one-third fewer parameters, though its three sweeps require more inference compute.
Thus, the perplexity gains carry over to downstream accuracy.

\vspace{-2mm}
\subsection{{Composing with Residual-Stream Enhancements}}
\label{sec:residual_experiments}

{The Lattice adds a horizontal pathway that carries latents across time,
whereas recent residual-stream enhancements reshape how information flows vertically,
across depth.
We test whether the two are complementary by pretraining 400M models on 20B FineWeb-Edu
tokens with hyper-connections (HC)~\citep{zhu2025hyperconnections}, manifold-constrained
hyper-connections (mHC)~\citep{xie2026mhcmanifoldconstrainedhyperconnections}, value
residuals~\citep{zhou2025valueresiduallearning}, and nanochat-style input residuals
(x0)~\citep{nanochat}, alone and in combination, with and without the Lattice
(Fig.~\ref{fig:residual_stream_comparison}); the Lattice models narrow the vertical width from
$1024$ to $932$ to remain iso-parameter (details in Appendix Sec.~\ref{app:training_residual}).
On its own, the Lattice lowers FineWeb-Edu perplexity from $15.41$ to $14.47$, more than HC
($15.28$), mHC ($14.93$), or value residuals ($14.69$), and is second only to x0 ($14.34$).
More importantly, its gains add to theirs: the Lattice lowers HC from $15.28$ to $14.21$
($-7.0\%$) and HC\,+\,x0 from $14.22$ to $13.61$ ($-4.3\%$).
\textit{Lattice\,+\,HC\,+\,x0 is the best model overall, $11.7\%$ below the plain Transformer and
$3.9\%$ below the strongest Transformer composition (x0\,+\,value, $14.16$).}
Thus, horizontal latents are not a substitute for a better residual stream; they are an
orthogonal axis that compounds with it.}

\vspace{-2mm}
\subsection{Recurrent Backbones}
\label{sec:recurrent_backbone_experiments}

We pretrain GDN, Mamba2, Triadic GDN, and GDN--Mamba2 hybrids without sliding-window
attention (no-SWA) on 10B FineWeb-Edu tokens, pairing each pure backbone with a Lattice
counterpart of the same native-state geometry and approximately matched parameter count
($\sim$405M).
We sweep 16K, 32K, 64K, 128K, and 256K native-state entries per mixer
(1K $=1024$ scalar entries), excluding the added horizontal latents and auxiliary
convolution states.
For GDN, the state size is $H d_k d_v$ for $H$ heads with per-head key and value
dimensions $d_k,d_v$; we use $(H,d_k,d_v)=(16,16,64)$, $(16,32,64)$, $(16,64,64)$,
$(8,128,128)$, and $(4,256,256)$, and vary the other backbones similarly (Appendix).
The horizontal latent modulates only the mixer's memory read, write, and update
quantities, not the MLP or decoding.
Perplexity results use \texttt{input-injection}\,+\,\texttt{vstream}, with the Lattice model's
vertical residual stream ($d_{\mathrm{model}}$) narrowed to offset the added parameters{, keeping every comparison iso-parameter}.
For downstream results, at each state budget and shot setting we select, by nine-task
average, the best of three couplings---\texttt{input-injection}\,+\,\texttt{vstream} and two
\texttt{vstream}-only variants that restore the parameter budget by reducing either
$d_{\mathrm{model}}$ or the FFN width (more details in Appendix Sec.~\ref{sec:appendix_experiments}).

\vspace{-2mm}
\paragraph{Language modeling.}
On FineWeb-Edu (Fig.~\ref{fig:recurrent_backbone_fineweb_results}), the Lattice lowers
perplexity at every native-state size: by $1.2$--$2.6\%$ for GDN, $0.6$--$2.2\%$ for Mamba2,
$1.1$--$2.1\%$ for Triadic GDN, and $0.3$--$1.0\%$ for the GDN--Mamba2 hybrid ($15.85$ vs.\
$15.92$ at 256K). {Lattice models can also outperform} pure models with four times as many native-state entries
per mixer: 64K Lattice GDN beats 256K pure GDN ($16.05$ vs.\ $16.11$), and 64K Lattice
Triadic GDN beats 256K pure Triadic GDN ($15.98$ vs.\ $16.09$).
These results are consistent with the hypothesis that horizontal latent helps vertical residual stream control memory writes and reads, enabling more effective use of native memory {(Sec.~\ref{sec:analysis})}.

\vspace{-2mm}
\paragraph{Downstream evaluation.}
On the same nine-task average (Figs.~\ref{fig:recurrent_backbone_downstream_0shot}
and~\ref{fig:recurrent_backbone_downstream_5shot}), the selected Lattice model outperforms
its pure counterpart in almost all paired comparisons, spanning three backbones (GDN,
Mamba2, and the no-SWA hybrid), five state budgets, and two shot settings.
Averaged over state budgets, the gains range from $+0.53$ to $+0.67$ points zero-shot and
$+0.76$ to $+0.95$ points five-shot across backbones, peaking at $+1.50$ (Mamba2, 16K,
five-shot), and they persist at the largest budget: at 256K, the Lattice adds $+0.86$
and $+1.06$ points to GDN. {Lattice models can again outperform} pure models with larger native states.
With $8\times$ fewer native entries per mixer, 32K Lattice GDN surpasses 256K pure GDN in
both settings ($44.88$ vs.\ $44.45$ zero-shot), as does the
32K hybrid over the 256K hybrid, and 16K Lattice Mamba2 over 128K pure Mamba2; with
$16\times$ fewer, 16K Lattice GDN still exceeds 256K pure GDN zero-shot ($44.57$ vs.\ $44.45$).

\vspace{-1mm}
\subsection{Analysis}
\label{sec:analysis}

\begin{wrapfigure}{r}{0.55\columnwidth}
    \centering
    \vspace{-8mm}
    \hspace{-4mm}
    \includegraphics[width=\linewidth]{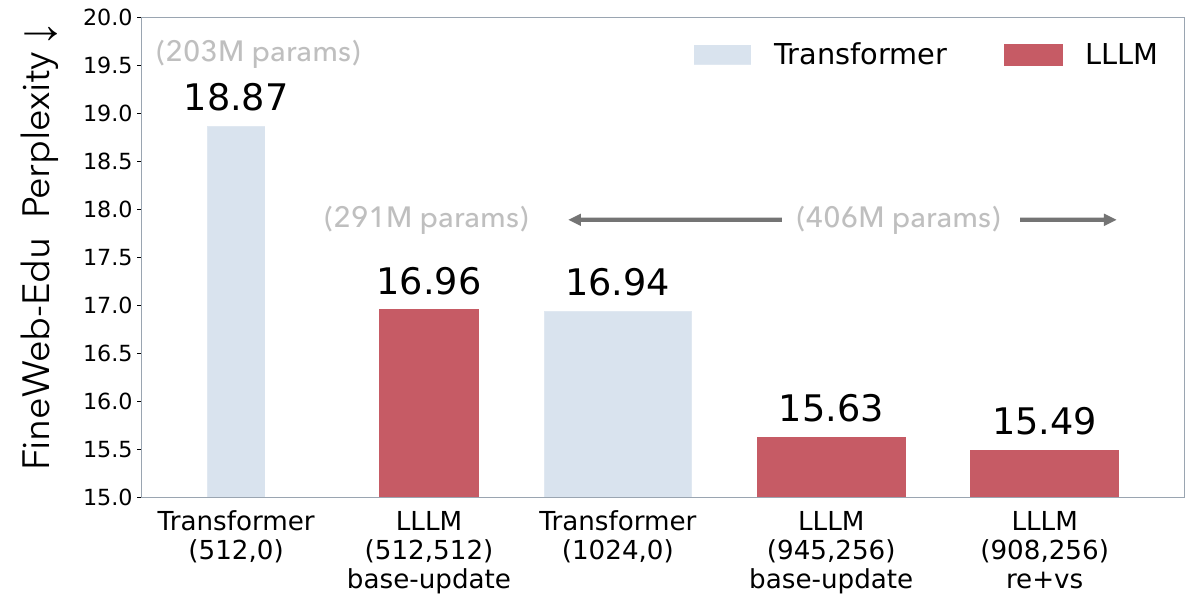}
    \vspace{-2mm}
    \caption{{\textbf{Allocating vertical and horizontal latents.}} Comparing {Transformers with LLLMs at different allocations} of vertical {and} horizontal latent {widths $(d_{\mathrm{model}},d_r)$ after 10B-token pretraining. See Sec.~\ref{sec:deconstructing_lattice}.}}
    \label{fig:dl_ablation_1}
\end{wrapfigure}

\vspace{-2mm}
\paragraph{Deconstructing the Lattice.}
\label{sec:deconstructing_lattice}
In this section, we ask where the gains come from, training all models on 10B FineWeb-Edu tokens \& reporting validation perplexity.
Narrowing the vertical latent alone is costly: Transformer$(512,0)$ reaches $18.87$, versus
$16.94$ for the 406\,M, 1024-wide Transformer (Fig.~\ref{fig:dl_ablation_1}).
Reinvesting that capacity horizontally more than recovers the loss: LLLM$(512,512)$ matches
the 406\,M Transformer with only 291\,M parameters ($16.96$), and at a matched 406\,M,
LLLM$(945,256)$ reaches $15.63$ ($-7.7\%$), or $15.49$ ($-8.5\%$) with
\texttt{input-injection}\,+\,\texttt{vstream}.
As shown in Fig.~\ref{fig:dl_ablation_3}, spending 200\,M more parameters on Transformer width ($15.98$) or depth ($15.82$) still
trails this 406\,M Lattice model.
Fig.~\ref{fig:dl_ablation_3} also isolates the top-to-bottom Jacobi refiner, which
feeds the top Lattice layer's horizontal latent to the bottom layer at the next token:
each J3 model (trained with three sweeps) improves on its base-update counterpart, e.g.,
from $14.83$--$14.89$ to $14.29$--$14.40$ at $\sim$600\,M, at additional inference compute.
Prior work~\citep{ponderlm2} has shown that training with a randomized number of Jacobi
iterations removes this additional inference cost, matching the performance of sequential rollout.

Figure~\ref{fig:dl_ablation_2} examines the mechanisms in the hardest regime, a
highly compressed 256-dimensional vertical latent ($d_{\mathrm{model}}=256$).
Here, adding horizontal capacity with the base update alone is not enough:
LLLM$(256,768)$ reaches $19.58$, worse than the 512-wide Transformer ($18.87$).
Jacobi refinement alone reduces this to $18.95$.
Separately, cumulatively adding input injection ($18.55$), \texttt{vstream} ($17.96$), the
horizontal decoding head with $d_s=512$ ($17.62$), a larger Lattice block (E4/L16/D4 to
E1/L20/D1; $17.42$), and $d_s=1024$ ($17.18$) closes most of the $2.64$-perplexity gap,
leaving the model $0.24$ behind the 406.7\,M Transformer with $21\%$ fewer parameters.
These additions also grow the model from 215.8\,M to 320.9\,M parameters, so they measure
combined gains rather than effects at fixed capacity.
LLLM$(512,512)$ still does better ($16.96$ with 291\,M parameters): these pathways mitigate
but do not remove the cost of an extremely narrow vertical latent, so the split
between vertical and horizontal capacity matters; most of our main models keep a wide
vertical latent with a smaller horizontal one ($d_r=256$).

\begin{table}[t]
  \centering
  \scriptsize
  \setlength{\tabcolsep}{8pt}
  \renewcommand{\arraystretch}{1.10}
  \begin{tabular}{@{}lccc@{}}
    \toprule
    \textbf{Native state size}
      & \textbf{Entropy effective rank} $r_{\mathrm{ent}}$ ($\downarrow$)
      & \textbf{Stable rank } $r_{\mathrm{stable}}$ ($\downarrow$)
      & \textbf{Query alignment gain } $G_1$ ($\uparrow$)\\
    \midrule
    16K  & $6.250 \!\rightarrow\! 5.649$   & $1.518 \!\rightarrow\! 1.353$ & $3.693 \!\rightarrow\! 5.579$  \\
    32K  & $10.077 \!\rightarrow\! 8.403$  & $1.629 \!\rightarrow\! 1.440$ & $5.613 \!\rightarrow\! 7.493$  \\
    64K  & $14.498 \!\rightarrow\! 11.842$ & $1.798 \!\rightarrow\! 1.555$ & $6.269 \!\rightarrow\! 8.925$  \\
    128K & $22.965 \!\rightarrow\! 19.444$ & $2.055 \!\rightarrow\! 1.766$ & $7.637 \!\rightarrow\! 10.714$ \\
    256K & $37.915 \!\rightarrow\! 33.418$ & $2.504 \!\rightarrow\! 2.128$ & $7.957 \!\rightarrow\! 13.321$ \\
    \bottomrule
  \end{tabular}
  \vspace{-1mm}
  \caption{\textbf{Native-memory geometry and query alignment in GDN.}
    Each cell reports \textbf{pure $\rightarrow$ Lattice}{, at matched native-state size and approximately matched parameters.
    Lower ranks mean the memory is concentrated in fewer directions; higher $G_1$ means queries align more with the leading memory direction.
    The Lattice improves all three metrics at every state size.
    Values are averaged over all $22$ layers and all heads, $1024$ FineWeb-Edu sequences of $2048$ tokens, and $15$ positions per sequence just after a state update. See Sec.~\ref{sec:analysis}.}}
    \vspace{-4mm}
  \label{tab:gdn_state_geometry}
\end{table}

\vspace{-2mm}
\paragraph{Memory geometry and query alignment.}
Finally, we ask how the Lattice changes the way GDN uses its native memory.
Each GDN head maintains a memory matrix $S\in\mathbb{R}^{d_k\times d_v}$, written with keys and
read with queries, where $d_k$ and $d_v$ are the per-head key and value dimensions.
Writing $S=U\Sigma V^\top$ with singular values $\sigma_1\ge\sigma_2\ge\cdots$ and leading left singular vector $u_1\in\mathbb{R}^{d_k}$, we measure how
concentrated the memory is, using effective and stable ranks.
The entropy effective rank $r_{\mathrm{ent}}=\exp\!\big(-\sum_i p_i\log p_i\big)$, with
$p_i=\sigma_i/\sum_j\sigma_j$, is the effective number of directions the spectrum spreads
over; the stable rank $r_{\mathrm{stable}}=\|S\|_F^2/\sigma_1^2=\sum_i\sigma_i^2/\sigma_1^2$
measures how strongly the leading direction dominates.
Both lie in $[1,\min(d_k,d_v)]$; lower values mean a more concentrated spectrum, but do not by
themselves imply less stored information.
We also measure the
query alignment gain $G_1=d_k\,(u_1^\top q)^2/\|q\|_2^2$: the squared cosine between the
normalized query $q\in\mathbb{R}^{d_k}$ used for reading and $u_1$,
scaled by $d_k$ so that an isotropic random query gives $\mathbb{E}[G_1]=1$.


Table~\ref{tab:gdn_state_geometry} shows a consistent pattern at every state budget:
the Lattice lowers $r_{\mathrm{ent}}$ by $9.6$--$18.3\%$ and $r_{\mathrm{stable}}$ by
$10.9$--$15.0\%$, while raising $G_1$ by $33.5$--$67.4\%$.
The alignment gap widens with state size, from $+1.89$ at 16K to $+5.36$ at 256K, where
Lattice queries align with the leading memory direction $13.3\times$ more than chance,
versus $8.0\times$ for the pure model.
Thus, with Lattice, GDN \textit{writes a more concentrated memory and reads more strongly
along its dominant direction}, consistent with the hypothesis that the horizontal latent
helps the vertical stream control memory writes and reads.
 
\vspace{-2mm}
\section{Conclusion}

\vspace{-2mm}
We introduced the \emph{Lattice of Latents}, which distributes computational
capacity across the vertical residual stream and persistent, layer-wise
horizontal latents.
Rather than replacing sequence memory, these latents act as companion
computational states that steer the backbone's computation.
We use an affine horizontal update and introduce enhancements that increase
latent expressivity while preserving token-parallel teacher-forced computation
within each pass.
LLLMs improve language modeling and downstream task averages across Transformer
and recurrent backbones.
The Lattice also composes with residual-stream enhancements:
combined with hyper-connections and input residuals, it outperforms every
Transformer variant we test.
On recurrent backbones, gains arise at matched native-state sizes and
approximately matched parameter counts,
with Lattice GDN and Lattice Triadic GDN outperforming
pure baselines that use $4\times$ as many native-state entries.
Overall, our results suggest that distributing active computational capacity
across interacting latents is a promising complement to scaling width,
depth, memory, or the residual stream alone.

\section*{Acknowledgements}
\vspace{-2mm}
We thank Aman Dudeja, David Jin, Jyo Pari, Han Guo, Jayden Teoh, Oliver Sieberling, Rachit Saxena, Seungwook Han, Shamit Lal, and Xingjian Bai for
insightful discussions and constructive feedback.
Shivam Duggal is particularly grateful to Prof.\ Phillip Isola for
feedback on the manuscript and valuable guidance over years that
helped inspire ideas in this work.
Antonio Torralba acknowledges support from ONR MURI grant
\#037564-00001.
Shivam Duggal additionally acknowledges support from the
Amazon AI Research Innovation Fellowship.

\bibliography{iclr2027_conference}

@misc{ouro,
      title={Scaling Latent Reasoning via Looped Language Models}, 
      author={Rui-Jie Zhu and Zixuan Wang and Kai Hua and Tianyu Zhang and Ziniu Li and Haoran Que and Boyi Wei and Zixin Wen and Fan Yin and He Xing and Lu Li and Jiajun Shi and Kaijing Ma and Shanda Li and Taylor Kergan and Andrew Smith and Xingwei Qu and Mude Hui and Bohong Wu and Qiyang Min and Hongzhi Huang and Xun Zhou and Wei Ye and Jiaheng Liu and Jian Yang and Yunfeng Shi and Chenghua Lin and Enduo Zhao and Tianle Cai and Ge Zhang and Wenhao Huang and Yoshua Bengio and Jason Eshraghian},
      year={2026},
      eprint={2510.25741},
      archivePrefix={arXiv},
      primaryClass={cs.CL},
      url={https://arxiv.org/abs/2510.25741}, 
}

@misc{melt,
      title={Memory-Efficient Looped Transformer: Decoupling Compute from Memory in Looped Language Models}, 
      author={Victor Conchello Vendrell and Arnau Padres Masdemont and Niccolò Grillo and Jordi Ros-Giralt and Arash Behboodi and Fabio Valerio Massoli},
      year={2026},
      eprint={2605.07721},
      archivePrefix={arXiv},
      primaryClass={cs.CL},
      url={https://arxiv.org/abs/2605.07721}, 
}

@misc{neill2026loopedlatentattentioncrossloop,
      title={Looped Latent Attention: Cross-Loop KV Compression for Looped Transformers}, 
      author={James O' Neill and Fergal Reid},
      year={2026},
      eprint={2607.15456},
      archivePrefix={arXiv},
      primaryClass={cs.LG},
      url={https://arxiv.org/abs/2607.15456}, 
}

@article{coconut,
  title={Training Large Language Models to Reason in a Continuous Latent Space},
  author={Hao, Shibo and Sukhbaatar, Sainbayar and Su, DiJia and Li, Xian and Hu, Zhiting and Weston, Jason and Tian, Yuandong},
  journal={arXiv preprint arXiv:2412.06769},
  year={2024}
}

@misc{CODI,
  title={CODI: Compressing Chain-of-Thought into Continuous Space via Self-Distillation}, 
  author={Zhenyi Shen and Hanqi Yan and Linhai Zhang and Zhanghao Hu and Yali Du and Yulan He},
  year={2025},
  eprint={2502.21074},
  archivePrefix={arXiv},
  primaryClass={cs.CL},
  url={https://arxiv.org/abs/2502.21074}, 
}

@misc{ponderlm2,
      title={PonderLM-2: Pretraining LLM with Latent Thoughts in Continuous Space}, 
      author={Boyi Zeng and He Li and Shixiang Song and Yixuan Wang and Zitong Wang and Ziwei He and Xinbing Wang and Zhouhan Lin},
      year={2026},
      eprint={2509.23184},
      archivePrefix={arXiv},
      primaryClass={cs.CL},
      url={https://arxiv.org/abs/2509.23184}, 
}

@misc{gu2024mambalineartimesequencemodeling,
      title={Mamba: Linear-Time Sequence Modeling with Selective State Spaces}, 
      author={Albert Gu and Tri Dao},
      year={2024},
      eprint={2312.00752},
      archivePrefix={arXiv},
      primaryClass={cs.LG},
      url={https://arxiv.org/abs/2312.00752}, 
}

@misc{dao2024transformersssmsgeneralizedmodels,
      title={Transformers are SSMs: Generalized Models and Efficient Algorithms Through Structured State Space Duality}, 
      author={Tri Dao and Albert Gu},
      year={2024},
      eprint={2405.21060},
      archivePrefix={arXiv},
      primaryClass={cs.LG},
      url={https://arxiv.org/abs/2405.21060}, 
}

@misc{yang2025gateddeltanetworksimproving,
      title={Gated Delta Networks: Improving Mamba2 with Delta Rule}, 
      author={Songlin Yang and Jan Kautz and Ali Hatamizadeh},
      year={2025},
      eprint={2412.06464},
      archivePrefix={arXiv},
      primaryClass={cs.CL},
      url={https://arxiv.org/abs/2412.06464}, 
}

@misc{deltaproduct,
      title={DeltaProduct: Improving State-Tracking in Linear RNNs via Householder Products}, 
      author={Julien Siems and Timur Carstensen and Arber Zela and Frank Hutter and Massimiliano Pontil and Riccardo Grazzi},
      year={2025},
      eprint={2502.10297},
      archivePrefix={arXiv},
      primaryClass={cs.LG},
      url={https://arxiv.org/abs/2502.10297}, 
}

@misc{m2rnn,
      title={M$^2$RNN: Non-Linear RNNs with Matrix-Valued States for Scalable Language Modeling}, 
      author={Mayank Mishra and Shawn Tan and Ion Stoica and Joseph Gonzalez and Tri Dao},
      year={2026},
      eprint={2603.14360},
      archivePrefix={arXiv},
      primaryClass={cs.LG},
      url={https://arxiv.org/abs/2603.14360}, 
}

@misc{danieli2025pararnnunlockingparalleltraining,
      title={ParaRNN: Unlocking Parallel Training of Nonlinear RNNs for Large Language Models}, 
      author={Federico Danieli and Pau Rodriguez and Miguel Sarabia and Xavier Suau and Luca Zappella},
      year={2025},
      eprint={2510.21450},
      archivePrefix={arXiv},
      primaryClass={cs.LG},
      url={https://arxiv.org/abs/2510.21450}, 
}

@misc{pccot,
      title={Parallel Continuous Chain-of-Thought with Jacobi Iteration}, 
      author={Haoyi Wu and Zhihao Teng and Kewei Tu},
      year={2026},
      eprint={2506.18582},
      archivePrefix={arXiv},
      primaryClass={cs.CL},
      url={https://arxiv.org/abs/2506.18582}, 
}

@misc{perez2017filmvisualreasoninggeneral,
      title={FiLM: Visual Reasoning with a General Conditioning Layer}, 
      author={Ethan Perez and Florian Strub and Harm de Vries and Vincent Dumoulin and Aaron Courville},
      year={2017},
      eprint={1709.07871},
      archivePrefix={arXiv},
      primaryClass={cs.CV},
      url={https://arxiv.org/abs/1709.07871}, 
}

@misc{dehghani2019universaltransformers,
      title={Universal Transformers}, 
      author={Mostafa Dehghani and Stephan Gouws and Oriol Vinyals and Jakob Uszkoreit and Łukasz Kaiser},
      year={2019},
      eprint={1807.03819},
      archivePrefix={arXiv},
      primaryClass={cs.CL},
      url={https://arxiv.org/abs/1807.03819}, 
}

@misc{geiping2025scalingtesttimecomputelatent,
      title={Scaling up Test-Time Compute with Latent Reasoning: A Recurrent Depth Approach}, 
      author={Jonas Geiping and Sean McLeish and Neel Jain and John Kirchenbauer and Siddharth Singh and Brian R. Bartoldson and Bhavya Kailkhura and Abhinav Bhatele and Tom Goldstein},
      year={2025},
      eprint={2502.05171},
      archivePrefix={arXiv},
      primaryClass={cs.LG},
      url={https://arxiv.org/abs/2502.05171}, 
}

@misc{lieber2024jambahybridtransformermambalanguage,
      title={Jamba: A Hybrid Transformer-Mamba Language Model}, 
      author={Opher Lieber and Barak Lenz and Hofit Bata and Gal Cohen and Jhonathan Osin and Itay Dalmedigos and Erez Safahi and Shaked Meirom and Yonatan Belinkov and Shai Shalev-Shwartz and Omri Abend and Raz Alon and Tomer Asida and Amir Bergman and Roman Glozman and Michael Gokhman and Avashalom Manevich and Nir Ratner and Noam Rozen and Erez Shwartz and Mor Zusman and Yoav Shoham},
      year={2024},
      eprint={2403.19887},
      archivePrefix={arXiv},
      primaryClass={cs.CL},
      url={https://arxiv.org/abs/2403.19887}, 
}

@misc{dong2024hymbahybridheadarchitecturesmall,
      title={Hymba: A Hybrid-head Architecture for Small Language Models}, 
      author={Xin Dong and Yonggan Fu and Shizhe Diao and Wonmin Byeon and Zijia Chen and Ameya Sunil Mahabaleshwarkar and Shih-Yang Liu and Matthijs Van Keirsbilck and Min-Hung Chen and Yoshi Suhara and Yingyan Lin and Jan Kautz and Pavlo Molchanov},
      year={2024},
      eprint={2411.13676},
      archivePrefix={arXiv},
      primaryClass={cs.CL},
      url={https://arxiv.org/abs/2411.13676}, 
}

@misc{wang2026fullbandwidthtransformer,
      title={Full-bandwidth transformer}, 
      author={Xi Wang and Ziyang Cai and Zheng Zhan and Harry Dong and Ying Fan and Gustavo de Rosa and Tim Pearce and John Langford},
      year={2026},
      eprint={2608.08888},
      archivePrefix={arXiv},
      primaryClass={cs.AI},
      url={https://arxiv.org/abs/2608.08888}, 
}

@misc{huang2026latentrecurrenttransformerarchitecture,
      title={Latent Recurrent Transformer: Architecture Exploration, Training Strategies, and Scaling Behavior}, 
      author={Zeyi Huang and Xuehai He and LiLiang Ren and Yiping Wang and Baolin Peng and Hao Cheng and Shuohang Wang and Pengcheng He and Jianfeng Gao and Yong Jae Lee and Yelong Shen},
      year={2026},
      eprint={2605.26797},
      archivePrefix={arXiv},
      primaryClass={cs.LG},
      url={https://arxiv.org/abs/2605.26797}, 
}

@inproceedings{
    cai2026t2mlrtransformertemporalmiddlelayer,
    title={T2{MLR}: Transformer with Temporal Middle-Layer Recurrence},
    author={Ziyang Cai and Xingyu Zhu and Yihe Dong and Yinghui He and Sanjeev Arora},
    booktitle={Workshop on Latent {\&} Implicit Thinking {\textendash} Going Beyond CoT Reasoning},
    year={2026},
    url={https://openreview.net/forum?id=fQbk1EQWBO}
}

@misc{monea2026statepredictionseparationhypothesis,
      title={The State-Prediction Separation Hypothesis}, 
      author={Giovanni Monea and Nathan Godey and Kianté Brantley and Yoav Artzi},
      year={2026},
      eprint={2607.01218},
      archivePrefix={arXiv},
      primaryClass={cs.CL},
      url={https://arxiv.org/abs/2607.01218}, 
}

@misc{mozer2026recirculation,
      title={Recirculation}, 
      author={Michael C. Mozer and Shoaib Ahmed Siddiqui and Danny Sawyer and Sunny Sanyal and Rosanne Liu},
      year={2026},
      eprint={2608.17981},
      archivePrefix={arXiv},
      primaryClass={cs.LG},
      url={https://arxiv.org/abs/2608.17981}, 
}

@misc{zhu2025hyperconnections,
      title={Hyper-Connections}, 
      author={Defa Zhu and Hongzhi Huang and Zihao Huang and Yutao Zeng and Yunyao Mao and Banggu Wu and Qiyang Min and Xun Zhou},
      year={2025},
      eprint={2409.19606},
      archivePrefix={arXiv},
      primaryClass={cs.LG},
      url={https://arxiv.org/abs/2409.19606}, 
}

@misc{xie2026mhcmanifoldconstrainedhyperconnections,
      title={mHC: Manifold-Constrained Hyper-Connections}, 
      author={Zhenda Xie and Yixuan Wei and Huanqi Cao and Chenggang Zhao and Chengqi Deng and Jiashi Li and Damai Dai and Huazuo Gao and Jiang Chang and Kuai Yu and Liang Zhao and Shangyan Zhou and Zhean Xu and Zhengyan Zhang and Wangding Zeng and Shengding Hu and Yuqing Wang and Jingyang Yuan and Lean Wang and Wenfeng Liang},
      year={2026},
      eprint={2512.24880},
      archivePrefix={arXiv},
      primaryClass={cs.CL},
      url={https://arxiv.org/abs/2512.24880}, 
}

@misc{kimiteam2026attentionresiduals,
      title={Attention Residuals}, 
      author={Kimi Team and Guangyu Chen and Yu Zhang and Jianlin Su and Weixin Xu and Siyuan Pan and Yaoyu Wang and Yucheng Wang and Guanduo Chen and Bohong Yin and Yutian Chen and Junjie Yan and Ming Wei and Y. Zhang and Fanqing Meng and Chao Hong and Xiaotong Xie and Shaowei Liu and Enzhe Lu and Yunpeng Tai and Yanru Chen and Xin Men and Haiqing Guo and Y. Charles and Haoyu Lu and Lin Sui and Jinguo Zhu and Zaida Zhou and Weiran He and Weixiao Huang and Xinran Xu and Yuzhi Wang and Guokun Lai and Yulun Du and Yuxin Wu and Zhilin Yang and Xinyu Zhou},
      year={2026},
      eprint={2603.15031},
      archivePrefix={arXiv},
      primaryClass={cs.CL},
      url={https://arxiv.org/abs/2603.15031}, 
}

@misc{zhou2025valueresiduallearning,
      title={Value Residual Learning}, 
      author={Zhanchao Zhou and Tianyi Wu and Zhiyun Jiang and Fares Obeid and Zhenzhong Lan},
      year={2025},
      eprint={2410.17897},
      archivePrefix={arXiv},
      primaryClass={cs.CL},
      url={https://arxiv.org/abs/2410.17897}, 
}

@article{teoh2026nextlatentpredictiontransformerslearn,
  title={Next-latent prediction transformers learn compact world models},
  author={Teoh, Jayden and Tomar, Manan and Ahn, Kwangjun and Hu, Edward S. and Pearce, Tim and Sharma, Pratyusha and Krishnamurthy, Akshay and Islam, Riashat and Lamb, Alex and Langford, John},
  journal={arXiv preprint arXiv:2511.05963},
  year={2025}
}

@misc{nanochat,
  author = {Andrej Karpathy},
  title = {nanochat: The best ChatGPT that \$100 can buy},
  year = {2025},
  publisher = {GitHub},
  url = {https://github.com/karpathy/nanochat}
}

@misc{allenzhu2026physicslanguagemodels41,
      title={Physics of Language Models: Part 4.1, Architecture Design and the Magic of Canon Layers}, 
      author={Zeyuan Allen-Zhu},
      year={2026},
      eprint={2512.17351},
      archivePrefix={arXiv},
      primaryClass={cs.CL},
      url={https://arxiv.org/abs/2512.17351}, 
}

@misc{sieberling2026dynamicshortconvolutionsimprove,
      title={Dynamic Short Convolutions Improve Transformers}, 
      author={Oliver Sieberling and Bharat Runwal and Rameswar Panda and Yoon Kim},
      year={2026},
      eprint={2606.03825},
      archivePrefix={arXiv},
      primaryClass={cs.LG},
      url={https://arxiv.org/abs/2606.03825}, 
}

@misc{mozer2026topologicaltroubletransformers,
      title={The Topological Trouble With Transformers}, 
      author={Michael C. Mozer and Shoaib Ahmed Siddiqui and Rosanne Liu},
      year={2026},
      eprint={2604.17121},
      archivePrefix={arXiv},
      primaryClass={cs.LG},
      url={https://arxiv.org/abs/2604.17121}, 
}
\bibliographystyle{iclr2027_conference}

\newpage
\appendix
\section{Appendix}


\subsection{Extended Related Work}
\label{app:related_work}

\paragraph{More computation on a single stream.}
Extra computation along depth or time often comes with extra persistent memory.
In depth-based methods, each added layer
brings its own KV cache, and looped models may keep iteration-specific caches, although
recent work shares or compresses state across loops
\citep{melt, neill2026loopedlatentattentioncrossloop}.
Among continuous latent reasoning methods, Coconut feeds the final
hidden state back as the next input, CODI distills such reasoning from an explicit
chain-of-thought teacher, and PonderLM-2 brings it into pretraining \citep{coconut, CODI, ponderlm2};
when latent thoughts are held as extra positions, they also add KV-cache entries.
Because consecutive thoughts depend on each other, single-pass teacher forcing
no longer applies; PonderLM-2 and PCCoT
recover token parallelism with Jacobi-style refinement \citep{ponderlm2, pccot}, which
inspires our optional Jacobi refiner (App.~\ref{app:jacobi}).

\paragraph{Residual-stream enhancements.}
We detail three residual-stream enhancements: input residuals, value residuals, and
hyper-connections.
An input residual re-injects the normalized token embedding before every layer,
$h_t^{\ell-1}\leftarrow\lambda_r^\ell h_t^{\ell-1}+\lambda_0^\ell\,\mathrm{Norm}(h_t^0)$,
with learned per-layer scalars $\lambda_r^\ell,\lambda_0^\ell$.
Value residual learning \citep{zhou2025valueresiduallearning} adds a similar shortcut inside
attention: each layer mixes its attention values with those of the first layer,
$v_t^\ell\leftarrow\lambda_1^\ell v_t^1+\lambda_2^\ell v_t^\ell$, with learned per-layer scalars,
so later layers retain direct access to early token information.
Hyper-connections \citep{zhu2025hyperconnections} expand the residual stream into $n$
parallel streams: each block reads a learned, input-dependent mixture of the streams,
and its output is written back as the streams are remixed; mHC
\citep{xie2026mhcmanifoldconstrainedhyperconnections} constrains this mixing to be
doubly stochastic.
All three act within a timestep, so they compose directly with the Lattice
(Sec.~\ref{sec:residual_compose}); the horizontal update reads whatever input the
residual design produces. With hyper-connections one horizontal latent per
layer is shared across the $n$ streams.

\paragraph{Recurrent sequence models.}
Beyond Mamba-2's gated recurrence, Gated DeltaNet adds delta-rule state updates, and DeltaProduct
composes several Householder-style corrections into a more expressive but still linear
transition \citep{yang2025gateddeltanetworksimproving, deltaproduct}.
Work on nonlinear recurrences is complementary: ParaRNN parallelizes general nonlinear
recurrences with Newton
iterations, and M$^2$RNNs combine nonlinear updates with matrix-valued states
\citep{danieli2025pararnnunlockingparalleltraining, m2rnn}.
We keep the temporal update affine, add expressivity through its inputs and readouts,
and obtain nonlinear feedback across time only through optional Jacobi refinement.
Advances in parallelizable nonlinear recurrences could further strengthen the
horizontal latent as a temporal companion to the residual stream.

\paragraph{Recurrence within Transformers (concurrent work).}
Full-Bandwidth Transformer \citep{wang2026fullbandwidthtransformer} feeds the previous
token's top-layer state back alongside the next embedding; Latent Recurrent Transformer
\citep{huang2026latentrecurrenttransformerarchitecture} reuses a selected high-level
representation as recurrent memory; and T$^2$MLR
\citep{cai2026t2mlrtransformertemporalmiddlelayer} routes a deeper representation from
the previous token into an earlier layer through a gated pathway.
T$^2$MLR is the closest to our Jacobi refiner, but its recurrent object is a
cached vertical representation, whereas our Jacobi refiner links the top and bottom of a
lattice of horizontal states that each follow their own temporal recurrence.
Recirculation \citep{mozer2026recirculation} also sends deep activations to shallower
layers across steps, but is a training-free, inference-time intervention with
sequential prefill, whereas the Lattice is trained end to end in a token-parallel fashion. The State-Prediction Separation Transformer \citep{monea2026statepredictionseparationhypothesis} shares our view that a single
representation can be burdened by multiple roles, but separates them within each token's computation rather than keeping a persistent latent at each layer.

\definecolor{LLLMBlockGray}{RGB}{245,245,245}

\begin{table*}[t]
  \centering

  \small
  \setlength{\tabcolsep}{4.0pt}
  \renewcommand{\arraystretch}{1.12}
  \sisetup{
    table-number-alignment = center,
    group-digits           = false
  }

  \resizebox{\textwidth}{!}{%
  \begin{tabular}{@{}
    l
    S[table-format=3.0]
    S[table-format=2.3]
    S[table-format=2.3]
    S[table-format=2.3]
    S[table-format=2.2]
    S[table-format=2.3]
    S[table-format=2.2]
  @{}}
    \toprule

    \multirow{2}{*}{\textbf{Approach}}
      & \multicolumn{1}{c}{
          \multirow{2}{*}{
            \shortstack{
              \textbf{Params}\\[-0.2ex]
              \textbf{(M)}
            }
          }
        }
      & \multicolumn{1}{c}{
          \multirow{2}{*}{
            \shortstack{
              \textbf{FineWeb-Edu}\\[-0.2ex]
              \textbf{PPL} $\downarrow$
            }
          }
        }
      & \multicolumn{1}{c}{
          \multirow{2}{*}{
            \shortstack{
              \textbf{Pile}\\[-0.2ex]
              \textbf{PPL} $\downarrow$
            }
          }
        }
      & \multicolumn{2}{c}{\textbf{LAMBADA OpenAI}}
      & \multicolumn{2}{c}{\textbf{LAMBADA Standard}} \\

    \cmidrule(lr){5-6}
    \cmidrule(l){7-8}

      & & & &
      {\textbf{PPL} $\downarrow$}
      & {\textbf{Acc.} $\uparrow$}
      & {\textbf{PPL} $\downarrow$}
      & {\textbf{Acc.} $\uparrow$} \\

    \midrule

    \rowcolor{LLLMBlockGray}
    \multicolumn{8}{@{}l@{}}{%
      \rule{0pt}{2.7ex}%
      \textbf{400M-parameter models}%
    } \\

    Base-24L (width $d_v=1024$)
      & 407
      & 14.719
      & 26.518
      & 40.759
      & 34.83
      & 69.346
      & 26.49 \\

    LLLM $(945,256)$
      & 407
      & 14.050
      & 25.081
      & 37.605
      & 37.36
      & 64.314
      & 28.80 \\

    LLLM $(908,256)$ re+vs
      & 406
      & \textbf{13.921}
      & \textbf{24.148}
      & \textbf{37.451}
      & \textbf{37.73}
      & \textbf{61.598}
      & \textbf{29.89} \\

    \midrule

    \rowcolor{LLLMBlockGray}
    \multicolumn{8}{@{}l@{}}{%
      \rule{0pt}{2.7ex}%
      \textbf{500M-parameter models}%
    } \\

    LLLM $(512,512)$, 40L
      & 495
      & 13.831
      & 24.203
      & 37.525
      & 39.38
      & 63.721
      & 30.84 \\

    \midrule

    \rowcolor{LLLMBlockGray}
    \multicolumn{8}{@{}l@{}}{%
      \rule{0pt}{2.7ex}%
      \textbf{600M-parameter models}%
    } \\

    Base-24L (width $d_v=1620$)
      & 603
      & 13.944
      & 26.793
      & 37.404
      & 37.53
      & 62.158
      & 28.31 \\

    Base-24L (width $d_v=1472$)
      & 606
      & 14.033
      & 25.016
      & 38.501
      & 36.79
      & 64.490
      & 29.94 \\

    Base-40L (width $d_v=1024$)
      & 609
      & 13.802
      & \textbf{23.478}
      & 38.123
      & 38.54
      & 63.146
      & 30.16 \\

    LLLM $(1024,1024)$
      & 603
      & \textbf{13.189}
      & 23.818
      & \textbf{36.365}
      & \textbf{40.33}
      & \textbf{58.516}
      & \textbf{30.91} \\

    \bottomrule
  \end{tabular}%
  }

  \caption{
    \textbf{Language-modeling quality after 30B-token FineWeb-Edu
    pretraining.}
    We report perplexity ($\downarrow$) on FineWeb-Edu, Pile, and the
    OpenAI and Standard LAMBADA variants, together with LAMBADA
    last-word accuracy ($\uparrow$). Models are grouped by parameter
    scale. The performance trends are similar to the 20b token results in Tab.~\ref{tab:20b_arch_fineweb_ood}.
  }
  \label{tab:30b_arch_fineweb_ood}
\end{table*}

\subsection{Additional Method Details}
\label{app:method_details}

\subsubsection{Decoding Modulation (How the horizontal latents act)}
\label{app:decoding_modulation}
When this optional head is enabled, the topmost Lattice layer's horizontal
latent, with its own normalization, joins the final vertical latent in a shared
readout space of dimension $d_s$, and the next token is decoded from their sum:
\begin{equation}
p(x_{t+1}\mid x_{\le t})=\mathrm{softmax}\Big(
W_U\big(P_v\mathrm{Norm}(h_t^N)
+P_r\mathrm{Norm}_{\mathrm{head}}(r_t^{\mathrm{top}})\big)\Big),
\notag
\end{equation}
where $P_v\in\mathbb{R}^{d_s\times d_{\mathrm{model}}}$ and
$P_r\in\mathbb{R}^{d_s\times d_r}$ are learned projections and
$W_U\in\mathbb{R}^{V\times d_s}$ maps to vocabulary logits.
Like every other injection in Sec.~\ref{sec:horizontal_act}, $P_r$ is zero-initialized.
Hence at initialization the horizontal latent contributes nothing: attention and
the MLP reduce to the standard Transformer layer, and the output is decoded from
the vertical latent alone. The horizontal latent gains influence only through
training.
We use decoding modulation only in experiments with a highly compressed
vertical latent ($d_{\mathrm{model}}{=}256$).
When enabled, the horizontal decoding head reads the separately normalized raw
top-layer state $r_t^{\mathrm{top}}$, not the \texttt{vstream}-enriched readout.

\subsubsection{Input Injection (How the horizontal latents are updated)}
\label{app:input_injection}
The base update receives new information only through $h_t^{\ell-1}$, so its
access to the current input is limited by what the vertical latent preserves, which we
probe by ablating a narrower vertical latent (Fig.~\ref{fig:dl_ablation_3}).
By layer $\ell$, the vertical latent may also have discarded information about the
raw token identity.
\textbf{Input injection} (\texttt{inj}) therefore feeds the token embedding, with its
own normalized read $\hat h_t^0=\mathrm{Norm}_e(h_t^0)$, directly into both
pre-activations of the update:
\begin{align*}
\Delta_t^\ell
&=\mathrm{softplus}\big(
W_\Delta^\ell n_t^\ell+W_{e,\Delta}^\ell\hat h_t^0+b_\Delta^\ell\big)\\
c_t^\ell
&=\mathrm{SiLU}\big(W_r^\ell n_t^\ell+W_{e,c}^\ell\hat h_t^0+b_r^\ell\big)
\end{align*}
with $W_{e,\Delta}^\ell$ and $W_{e,c}^\ell$ zero-initialized.
The recurrence $r_t^\ell=g_t^\ell\odot r_{t-1}^\ell+b_t^\ell$ is otherwise unchanged:
$\Delta_t^\ell$ and $c_t^\ell$ still depend only on current-timestep inputs
($n_t^\ell$ and $\hat h_t^0$), never on $r_{t-1}^\ell$, so the transition remains
exactly affine and the same parallel scan applies.

\subsubsection{Jacobi Refiner (How the horizontal latents are updated)}
\label{app:jacobi}

\begin{figure}[t]
    \centering
    \includegraphics[width=0.9\linewidth]{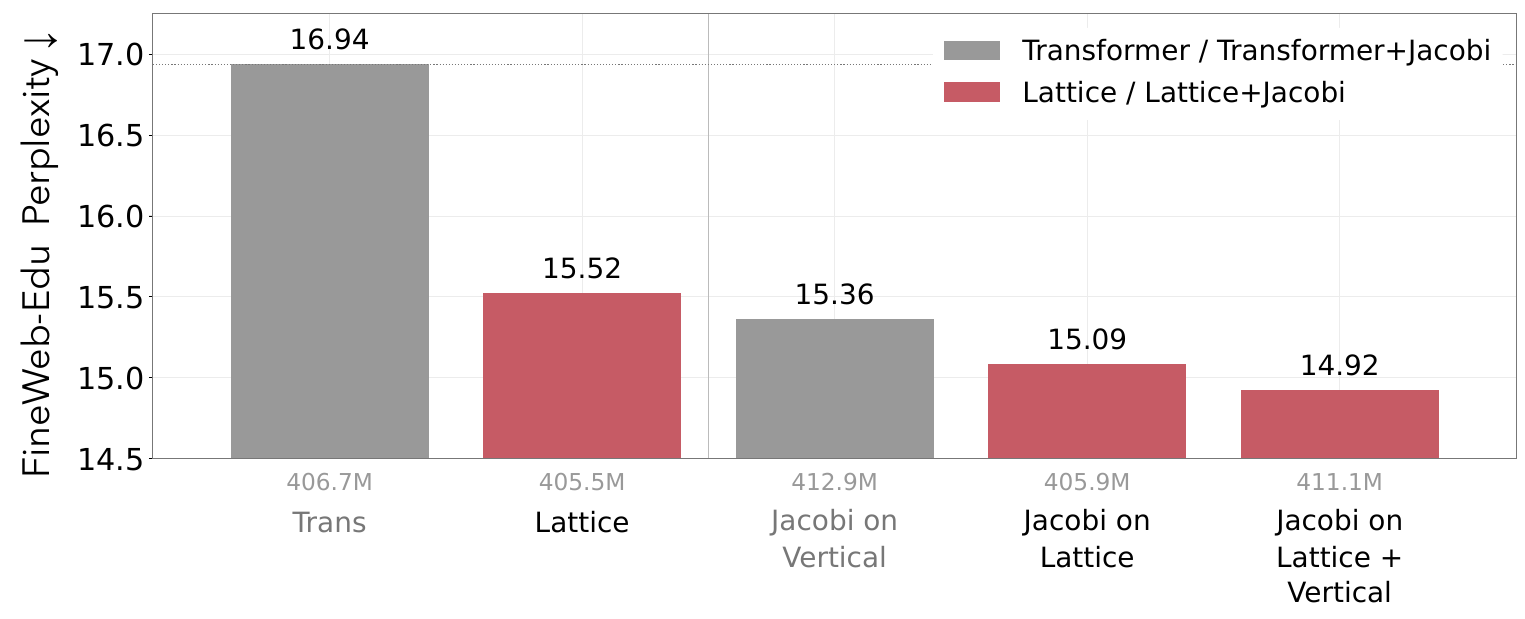}
    \vspace{-4mm}
    \caption{\textbf{Ablating top-to-bottom residual-stream flow.} {Feeding each token's top-layer state back into the bottom of the network at the next token adds a nonlinear path through the residual stream. \emph{Jacobi on vertical} applies this feedback to the vertical residual stream, analogous to the concurrent Full-Bandwidth Transformer~\citep{wang2026fullbandwidthtransformer}; \emph{Jacobi on Lattice} applies it to the horizontal latents; the last bar applies both. We call it ``Jacobi'' because the feedback is trained and evaluated with three parallel, teacher-forced Jacobi sweeps (J3; App.~\ref{app:jacobi}), at the cost of extra forward passes. Both forms of feedback help, and they compound: $16.94$ (Transformer) and $15.52$ (single-pass Lattice) fall to $14.92$ with feedback on both. FineWeb-Edu val perplexity, trained on 10B tokens.}}
    \label{fig:jacobi}
    \vspace{-4mm}
\end{figure}

The input-injection and \texttt{vstream} pathways use only information available at the
current timestep, so the base update keeps an exact parallel scan across time.
Neither feeds a \emph{higher} layer's horizontal latent back to a \emph{lower} layer at the
next timestep. Such top-to-bottom feedback is straightforward in a token-by-token rollout,
but breaks the single bottom-to-top sweep of teacher-forced training, because the top
layer's latent is not yet available when the bottom layer updates. Following
\citet{ponderlm2,pccot}, we approximate it during training with Jacobi iteration. Unlike
these works, which iterate over a recurrence that feeds the final \emph{vertical} latent
back as the next input, we apply it to the \emph{horizontal} latents.

\paragraph{Sweeps.}
Let $J\ge1$ be the total number of sweeps: \texttt{J0} denotes the base single pass and
\texttt{J3} a base pass followed by two refinement sweeps. Sweep $j=0$ runs the Lattice
block bottom to top exactly as in the base update, producing $r_t^{\ell,(0)}$ for every
layer and timestep. Each subsequent sweep $j=1,\dots,J-1$ takes the \emph{top} Lattice
layer's latent from the previous sweep, shifted one timestep back, and maps it through a
SwiGLU:
\begin{align*}
N_t^{(j)}
&=W_{J,\mathrm{down}}\big(
\mathrm{SiLU}(W_{J,\mathrm{gate}}\hat z_t^{(j-1)})
\odot(W_{J,\mathrm{up}}\hat z_t^{(j-1)})\big),
\qquad
\hat z_t^{(j-1)}=\mathrm{RMSNorm}\big(r_{t-1}^{\mathrm{top},(j-1)}\big),
\end{align*}
with hidden width $2d_r$, $\hat z$ set to zero at the first timestep, and $N_t^{(0)}=0$.
Only the bottom Lattice layer $\ell_0$ receives the correction, which is added to its
candidate content {{} $c_t^{\ell_0,(j)}=\mathrm{SiLU}(W_r^{\ell_0}n_t^{\ell_0,(j)}+b_r^{\ell_0})$,
computed from the normalized vertical input $n_t^{\ell_0,(j)}$ (plus the input-injection term when used;
full update in Sec.~\ref{sec:horizontal_update_sec})}:
\begin{align*}
b_t^{\ell_0,(j)}=(1-g_t^{\ell_0,(j)})\odot\big(c_t^{\ell_0,(j)}+N_t^{(j)}\big),
\end{align*}
which enters the same affine recurrence $r_t^\ell=g_t^\ell\odot r_{t-1}^\ell+b_t^\ell$.
Every sweep reruns the whole Lattice block from the same early-block outputs and the same
learned initial states, with parameters shared across sweeps and all layer inputs and
coefficients recomputed; the final sweep supplies the block output. Because $N_t^{(j)}$
depends only on the fully resolved previous sweep, each sweep's scan remains exactly affine
and token-parallel; the cost is $J$ serial sweeps instead of one.

\paragraph{Training and inference.}
$W_{J,\mathrm{gate}}$ and $W_{J,\mathrm{up}}$ are initialized from $\mathcal{N}(0,0.02^2)$
and $W_{J,\mathrm{down}}$ is zero-initialized, so a J3 model equals its J0 counterpart at
initialization. 
We train with a fixed number of
sweeps ($J=3$) and backpropagate through all of them. At inference, the top-layer latent
of the preceding token is already available, so exact recurrence
($N_t=F(r_{t-1}^{\mathrm{top}})$) can be evaluated in a single token-by-token pass without
refinement sweeps (``sequential rollout''); finite number of sweeps need not reproduce
this rollout. We report both three-sweep \& sequential-rollout eval
(Sec.~\ref{sec:transformer_backbones}).

\definecolor{ShotBlockGray}{RGB}{232,232,232}
\definecolor{LLLMBlockGray}{RGB}{245,245,245}

\begin{table*}[t]
  \centering

  \small
  \setlength{\tabcolsep}{3.8pt}
  \renewcommand{\arraystretch}{1.08}
  \sisetup{
    table-number-alignment = center,
    group-digits           = false
  }

  \resizebox{\textwidth}{!}{%
  \begin{tabular}{@{}
    l
    S[table-format=3.0]
    S[table-format=2.1]
    S[table-format=2.1]
    S[table-format=2.1]
    S[table-format=2.1]
    S[table-format=2.1]
    S[table-format=2.1]
    S[table-format=2.1]
    S[table-format=2.1]
    S[table-format=2.1]
    S[table-format=2.2]
  @{}}
    \toprule

    \multirow{2}{*}{\textbf{Approach}}
      & \multicolumn{1}{c}{
          \multirow{2}{*}{\textbf{Params (M)}}
        }
      & \multicolumn{2}{c}{\textbf{LAMBADA}}
      & \multicolumn{2}{c}{\textbf{ARC}}
      & \multicolumn{1}{c}{\multirow{2}{*}{\textbf{WG}}}
      & \multicolumn{1}{c}{\multirow{2}{*}{\textbf{PIQA}}}
      & \multicolumn{1}{c}{\multirow{2}{*}{\textbf{HS}}}
      & \multicolumn{1}{c}{\multirow{2}{*}{\textbf{SciQ}}}
      & \multicolumn{1}{c}{\multirow{2}{*}{\textbf{RACE}}}
      & \multicolumn{1}{c}{
          \multirow{2}{*}{\textbf{Avg.} $\uparrow$}
        } \\

    \cmidrule(lr){3-4}
    \cmidrule(lr){5-6}

      &
      & {\textbf{OpenAI}}
      & {\textbf{Standard}}
      & {\textbf{Easy}}
      & {\textbf{Challenge}}
      & & & & & & \\

    \midrule

    \rowcolor{ShotBlockGray}
    \multicolumn{12}{@{}c@{}}{%
      \rule{0pt}{3.0ex}%
      \textbf{0-shot accuracy}%
    } \\

    \rowcolor{LLLMBlockGray}
    \multicolumn{12}{@{}l@{}}{%
      \rule{0pt}{2.7ex}%
      \textbf{400M-parameter models}%
    } \\

    Base-24L (width $d_v=1024$)
      & 407
      & 33.0
      & 24.9
      & 53.7
      & 27.8
      & 50.6
      & 66.8
      & 40.2
      & 81.4
      & 30.7
      & 45.46 \\

    LLLM $(945,256)$
      & 407
      & \textbf{35.9}
      & 26.9
      & \textbf{56.4}
      & 29.0
      & \textbf{54.2}
      & \textbf{68.4}
      & 42.4
      & \textbf{84.4}
      & 31.3
      & \textbf{47.67} \\

    LLLM $(908,256)$ re+vs
      & 406
      & 35.1
      & \textbf{28.2}
      & 56.0
      & \textbf{29.6}
      & 51.6
      & 68.0
      & \textbf{43.6}
      & 82.6
      & \textbf{31.7}
      & 47.38 \\


    LLLM $(945,256)$ J3 
      & 407
      & \textbf{38.2}
      & \textbf{29.6}
      & \textbf{55.6}
      & \textbf{30.1}
      & \textbf{55.0}
      & \textbf{69.0}
      & \textbf{44.9}
      & \textbf{83.7}
      & \textbf{32.2}
      & \textbf{48.69} \\

    \midrule

    \rowcolor{LLLMBlockGray}
    \multicolumn{12}{@{}l@{}}{%
      \rule{0pt}{2.7ex}%
      \textbf{500M-parameter models}%
    } \\

    LLLM $(512,512)$, 40L
      & 495
      & 35.9
      & 28.9
      & 56.6
      & 27.8
      & 54.5
      & 68.9
      & 44.2
      & 83.4
      & 32.7
      & 48.10 \\

    \midrule

    \rowcolor{LLLMBlockGray}
    \multicolumn{12}{@{}l@{}}{%
      \rule{0pt}{2.7ex}%
      \textbf{600M-parameter models}%
    } \\

    Base-24L (width $d_v=1620$)
      & 603
      & 33.1
      & 25.5
      & 54.9
      & 28.8
      & 50.9
      & 66.7
      & 41.5
      & 82.9
      & 30.9
      & 46.14 \\

    Base-24L (width $d_v=1472$)
      & 606
      & 35.1
      & 27.1
      & 55.6
      & 28.8
      & 52.1
      & 68.0
      & 42.1
      & 84.3
      & 33.0
      & 47.36 \\

    Base-40L (width $d_v=1024$)
      & 609
      & 35.2
      & 27.4
      & 56.1
      & 29.1
      & \textbf{53.3}
      & 67.8
      & 42.6
      & 84.1
      & 30.5
      & 47.36 \\

    LLLM $(1024,1024)$
      & 603
      & \textbf{36.0}
      & \textbf{28.6}
      & \textbf{57.6}
      & \textbf{29.5}
      & 52.4
      & \textbf{69.1}
      & \textbf{45.7}
      & \textbf{84.5}
      & \textbf{33.3}
      & \textbf{48.53} \\

    \specialrule{0.12em}{0.8ex}{0.5ex}

    \rowcolor{ShotBlockGray}
    \multicolumn{12}{@{}c@{}}{%
      \rule{0pt}{3.0ex}%
      \textbf{5-shot accuracy}%
    } \\

    \rowcolor{LLLMBlockGray}
    \multicolumn{12}{@{}l@{}}{%
      \rule{0pt}{2.7ex}%
      \textbf{400M-parameter models}%
    } \\

    Base-24L (width $d_v=1024$)
      & 407
      & 25.5
      & 24.1
      & 59.3
      & \textbf{30.9}
      & 51.5
      & 66.8
      & 39.9
      & 86.5
      & 31.4
      & 46.19 \\

    LLLM $(945,256)$
      & 407
      & 27.3
      & 24.4
      & 63.6
      & 30.6
      & \textbf{52.2}
      & \textbf{67.7}
      & 42.6
      & \textbf{88.4}
      & \textbf{33.3}
      & 47.79 \\

    LLLM $(908,256)$ re+vs
      & 406
      & \textbf{29.8}
      & \textbf{26.9}
      & \textbf{65.0}
      & 30.8
      & 51.2
      & 67.5
      & \textbf{43.1}
      & 87.7
      & 31.4
      & \textbf{48.16} \\


    LLLM $(945,256)$ J3
      & 407
      & \textbf{29.2}
      & \textbf{26.3}
      & \textbf{65.2}
      & \textbf{33.5}
      & \textbf{54.3}
      & \textbf{68.9}
      & \textbf{45.3}
      & \textbf{90.2}
      & \textbf{31.5}
      & \textbf{49.38} \\

    \midrule

    \rowcolor{LLLMBlockGray}
    \multicolumn{12}{@{}l@{}}{%
      \rule{0pt}{2.7ex}%
      \textbf{500M-parameter models}%
    } \\

    LLLM $(512,512)$, 40L
      & 495
      & 29.1
      & 26.8
      & 64.1
      & 31.6
      & 52.3
      & 67.9
      & 43.8
      & 89.2
      & 31.7
      & 48.50 \\

    \midrule

    \rowcolor{LLLMBlockGray}
    \multicolumn{12}{@{}l@{}}{%
      \rule{0pt}{2.7ex}%
      \textbf{600M-parameter models}%
    } \\

    Base-24L (width $d_v=1620$)
      & 603
      & 24.4
      & 23.4
      & 62.0
      & 32.4
      & 51.3
      & 67.2
      & 41.6
      & 87.3
      & 30.0
      & 46.63 \\

    Base-24L (width $d_v=1472$)
      & 606
      & 26.4
      & 23.2
      & 62.8
      & 31.7
      & 52.0
      & \textbf{68.3}
      & 42.2
      & 89.2
      & 30.4
      & 47.34 \\

    Base-40L (width $d_v=1024$)
      & 609
      & 26.3
      & 24.1
      & 63.2
      & 31.6
      & 51.1
      & 67.4
      & 43.2
      & 88.6
      & 30.9
      & 47.37 \\

    LLLM $(1024,1024)$
      & 603
      & \textbf{29.3}
      & \textbf{28.2}
      & \textbf{65.6}
      & \textbf{32.9}
      & \textbf{52.8}
      & 68.0
      & \textbf{45.3}
      & \textbf{89.4}
      & \textbf{32.7}
      & \textbf{49.37} \\

    \bottomrule
  \end{tabular}%
  }

  \caption{
    \textbf{Zero- and five-shot downstream task accuracy at the
    20B-token FineWeb-Edu pretraining scale.}
    Models are grouped by parameter scale and shown in the same order
    under both evaluation settings. All task values are accuracies
    ($\%$; $\uparrow$). LAMBADA results are reported on the OpenAI and
    Standard variants, while ARC results are reported on ARC-Easy and
    ARC-Challenge. WG / HS denote WinoGrande / HellaSwag,
    respectively. Avg.\ is reported average across 9 tasks.
  }
  \vspace{-5mm}
  \label{tab:20b_downstream_zero_five_shot}
\end{table*}

\subsection{Additional Experimental Results}
\label{sec:appendix_experiments}

\subsubsection{Transformer Backbones}
\label{app:transformer_experiments}

\paragraph{Scaling to 30B tokens.}
Building on the 20B-token results in Table~\ref{tab:20b_arch_fineweb_ood},
Table~\ref{tab:30b_arch_fineweb_ood} shows that the benefits of the
Lattice persist under substantially longer, post-Chinchilla
pretraining on 30B tokens. At the matched 400M scale, LLLM$(908,256)$ with \texttt{input-injection}\,+\,\texttt{vstream} outperforms Base-24L on every reported
metric, reducing FineWeb-Edu, Pile, LAMBADA-OpenAI, and
LAMBADA-Standard perplexity by $5.4\%$, $8.9\%$, $8.1\%$, and
$11.2\%$, respectively, while improving the two LAMBADA accuracies by
$2.90$ and $3.40$ points. The gains also hold at 600M parameters:
LLLM$(1024,1024)$ improves over the similarly sized Base-40L by
$4.4\%$ on FineWeb-Edu, $4.6\%$ on LAMBADA-OpenAI, and $7.3\%$ on
LAMBADA-Standard, while increasing their accuracies by $1.79$ and
$0.75$ points; Base-40L remains modestly better on Pile. Thus, the
advantages observed at 20B tokens do not disappear with extended
pretraining. Moreover, the strong results obtained with different
vertical--horizontal allocations suggest that this capacity split is an
important tuning axis, and that a more systematic search may yield
further improvements.

\vspace{-2mm}
\paragraph{Downstream task performance.}
Table~\ref{tab:20b_downstream_zero_five_shot} breaks down the nine-task averages of
Sec.~\ref{sec:transformer_backbones} by task after 20B-token pretraining.
At the matched 400M scale, LLLM$(945,256)$ J3 improves on Base-24L on all nine tasks in both
the zero- and five-shot settings.
At 600M, LLLM$(1024,1024)$ improves on Base-40L on all nine tasks five-shot and on eight
of nine zero-shot, trailing only on WinoGrande ($52.4$ vs.\ $53.3$).
LLLM$(512,512)$ with 40 layers also exceeds the 609M Base-40L in both settings ($48.10$
vs.\ $47.36$ zero-shot; $48.50$ vs.\ $47.37$ five-shot) with 495M parameters.

\vspace{-2mm}
\paragraph{Top-to-bottom Residual Stream Refinement Ablations.}
{Fig.~\ref{fig:jacobi} tests whether top-to-bottom feedback across timesteps helps the vertical
residual stream, the horizontal latents, or both. All models are trained on 10B FineWeb-Edu tokens
with the Transformer recipe (App.~\ref{app:training_transformer}); Lattice models use the
\texttt{vstream}-only configuration ($d_{\mathrm{model}}{=}932$, $d_r{=}256$).
\emph{Jacobi on vertical} passes the previous token's final-layer state through a small MLP and adds
it to the next token's embedding, similar to the Full-Bandwidth
Transformer~\citep{wang2026fullbandwidthtransformer}. \emph{Jacobi on Lattice} applies the
Jacobi refiner (App.~\ref{app:jacobi}) to the horizontal latents, and the last variant applies
both. All three are trained and evaluated with three parallel, teacher-forced sweeps (J3).
Vertical feedback lowers the Transformer's perplexity from $16.94$ to $15.36$ ($-9.3\%$), slightly
below the single-pass Lattice ($15.52$), but it reruns the full network three times and adds
$6.3$\,M parameters. Feedback on horizontal latents lowers the Lattice from $15.52$ to $15.09$
($-2.8\%$). Combining both works the best, $14.92$
($-3.9\%$ relative to single-pass Lattice), so two forms of feedback are complementary.}

\begin{figure}[t]
    \centering
    \includegraphics[width=0.9\linewidth]{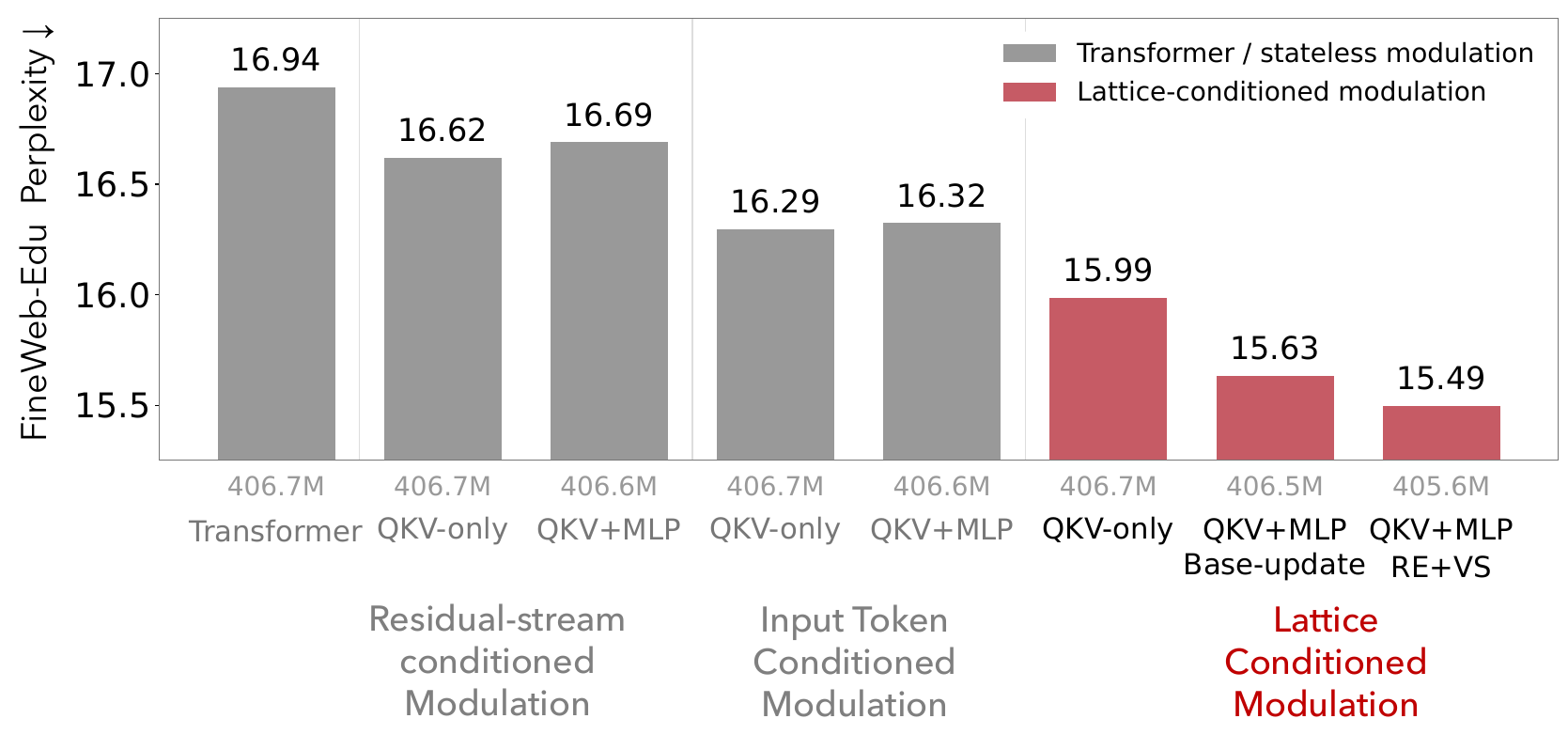}
    \vspace{-4mm}
    \caption{\textbf{Stateless vs.\ Lattice modulation: is recurrent history needed?} {We replace the horizontal latent in FiLM modulation with a stateless signal computed from either the token embedding (input-conditioned) or the incoming residual stream (residual-stream-conditioned), at matched $\sim$406M parameters. Lattice-conditioned modulation, which carries history across timesteps, clearly outperforms both stateless variants. FineWeb-Edu val perplexity, trained on 10B tokens.}}
    \label{fig:stateless_modulation}
    \vspace{-4mm}
\end{figure}

\vspace{-3mm}
\paragraph{Stateless vs.\ Lattice modulation.}
Is the gain of the Lattice due to its recurrent history, or would any extra input-dependent
FiLM conditioning suffice? Fig.~\ref{fig:stateless_modulation} replaces the horizontal latent
$\hat r_t^\ell$ with a \emph{stateless} signal
$z_t^\ell=\mathrm{Norm}\big(P^\ell\,\mathrm{Norm}(u_t)\big)\in\mathbb{R}^{256}$, where $P^\ell$ is a
learned per-layer projection and $u_t$ is either the layer's incoming residual stream $h_t^{\ell-1}$
(\emph{residual-stream-conditioned}) or the token embedding $h_t^0$ (\emph{input-conditioned}). We then perform FiLM based QKV and/or MLP modulation with this signal.
Everything else matches the Lattice: the same E1/L20/D1 split, the same zero-initialized FiLM heads on
QKV (and optionally the MLP), and widths adjusted so that all models have $\sim$406\,M parameters; only
the recurrence across timesteps is removed. All models are trained on 10B FineWeb-Edu tokens with the
Transformer recipe (App.~\ref{app:training_transformer}).
Stateless modulation helps: input-conditioned QKV modulation lowers the Transformer's perplexity
from $16.94$ to $16.29$ ($-3.8\%$), and conditioning on the raw embedding beats conditioning on the
residual stream ($16.62$), consistent with input injection helping the horizontal update.
Recurrent conditioning helps more at matched parameters: with QKV modulation alone the Lattice reaches
$15.99$ ($-1.9\%$ vs.\ the best stateless variant), and with QKV+MLP modulation $15.63$ ($-4.2\%$ vs.\
$16.32$), improving to $15.49$ with \texttt{inj}+\texttt{vs}. MLP modulation also helps only when the
conditioning carries history: it slightly hurts both stateless variants but improves the Lattice from
$15.99$ to $15.63$. These results are consistent with the hypothesis that the history accumulated in
the horizontal latents, rather than extra conditioning capacity alone, drives most of the Lattice's
gain.

\begin{figure}[t]
    \centering
    \includegraphics[width=0.9\linewidth]{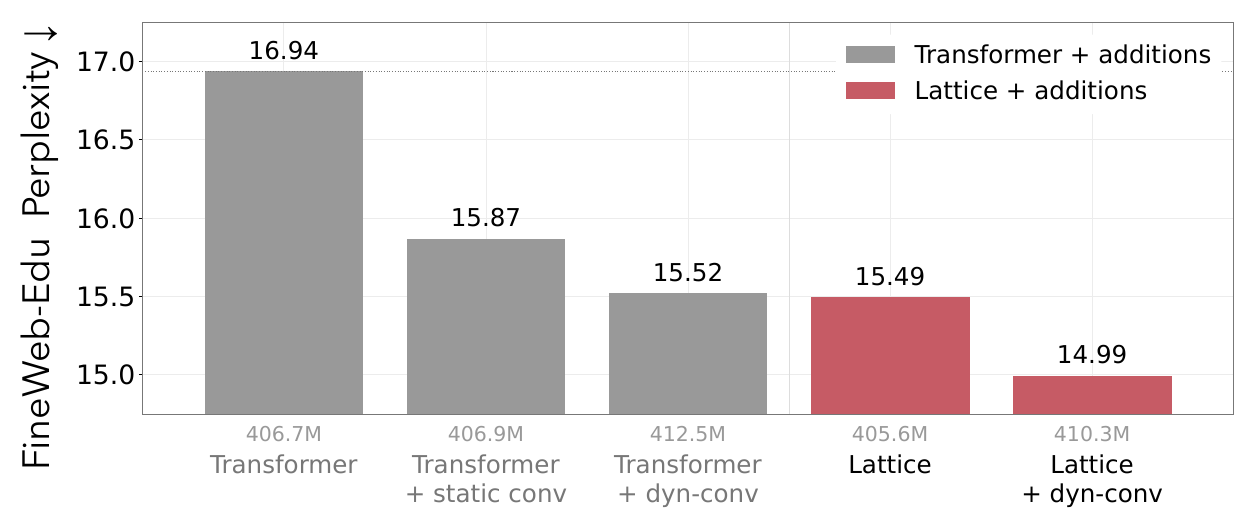}
    \vspace{-4mm}
    \caption{\textbf{Short-Conv vs.\ Lattice} Static and dynamic width-4 short convolutions on $q,k,v$ improve the Transformer; the Lattice matches dynamic convolution with fewer parameters, and the two compose: Lattice\,+\,dynamic convolution reaches $14.99$. Parameter counts are shown below the bars. FineWeb-Edu val perplexity, trained on 10B tokens. See App.~\ref{app:transformer_experiments}.}
    \label{fig:conv_comparison}
    \vspace{-4mm}
\end{figure}

\vspace{-3mm}
\paragraph{Lattice vs.\ short convolutions.}
Short causal convolutions on the attention inputs let local context shape $q,k,v$,
similar in spirit to how the Lattice conditions attention on history.
Static short convolutions, such as Canon layers \citep{allenzhu2026physicslanguagemodels41}, mix each token's
representation with those of its few preceding tokens using learned, input-independent filters;
dynamic short convolutions \citep{sieberling2026dynamicshortconvolutionsimprove} instead generate the filter
weights from the input.
Fig.~\ref{fig:conv_comparison} compares them after 10B-token pretraining.
A width-4 static depthwise convolution on $q,k,v$ adds only $0.28$\,M parameters to the
Transformer and lowers perplexity from $16.94$ to $15.87$ ($-6.3\%$); a width-4 dynamic
convolution reaches $15.52$ with $5.8$\,M extra parameters.
The Lattice 
matches dynamic convolution
($15.49$) with $6.8$\,M fewer parameters, and the two are complementary: adding dynamic convolution
to the Lattice lowers perplexity to $14.99$ ($-3.4\%$ vs.\ dynamic convolution alone).
A convolution, static or dynamic, aggregates only a fixed window of recent tokens, whereas the
horizontal latents carry history across the whole sequence through their recurrence, thus \textit{advances in
parallelizable nonlinear updates could strengthen the Lattice further}.

\vspace{-3mm}
\paragraph{GSM8K-Aug finetuning.}

We finetune each 20B-token checkpoint on GSM8K-Aug ($385{,}620$ problems), following
the chain-of-thought (CoT) stage of Coconut~\citep{coconut}: conditioned on the question, the model
is trained with the next-token loss on the intermediate equation steps and the final answer, with
question tokens masked from the loss.
{We sweep the learning rate separately for each model over
$\{10^{-4},4\times10^{-4},8\times10^{-4}\}$; $8\times10^{-4}$ performs best for every model.
All other settings are shared:} AdamW with {a} learning rate {that is} constant after a linear
warmup over the first $3\%$ of steps, wd $0.01$, gradient clipping at $1.0$, batch size
$128$, bf16, and $4$ epochs ($12$K steps), with one run (seed $0$) per model.
J3 models use 3 Jacobi sweeps in both finetuning \& decoding.

\definecolor{LLLMBlockGray}{RGB}{245,245,245}

\begin{wraptable}{r}{0.55\columnwidth}
  \centering
    \vspace{-4mm}
  \setlength{\abovecaptionskip}{0pt}
  \setlength{\belowcaptionskip}{4pt}

  \scriptsize
  \setlength{\tabcolsep}{1.2pt}
  \setlength{\fboxsep}{2pt}
  \renewcommand{\arraystretch}{1.10}

  \begin{tabularx}{\linewidth}{@{}
    >{\raggedright\arraybackslash}X
    >{\centering\arraybackslash}p{0.11\linewidth}
    >{\centering\arraybackslash}p{0.11\linewidth}
    >{\centering\arraybackslash}p{0.11\linewidth}
    >{\centering\arraybackslash}p{0.11\linewidth}
    >{\centering\arraybackslash}p{0.11\linewidth}
  @{}}
    \toprule

    \textbf{Approach}
      & 


      \shortstack{{GSM}\\[-0.2ex]{8K}}
      & \shortstack{{SVAMP}\\[-0.2ex]{1K}}
      & \shortstack{{Multi}\\[-0.2ex]{Arith}}
      & {ASDiv}
      & \shortstack{{GSM}\\[-0.2ex]{Hard}} \\

    \midrule

    \multicolumn{6}{@{}l@{}}{%
      \colorbox{LLLMBlockGray}{%
        \parbox{\dimexpr\linewidth-2\fboxsep\relax}{%
          \rule{0pt}{2.5ex}%
          \textbf{400M-parameter models}%
        }%
      }%
    } \\

    Base-24L ($d_v=1024$)
      & 29.7
      & 28.4
      & 74.0
      & 41.6
      & 6.4 \\

    LLLM $(945,256)$
      & 44.5
      & 42.0
      & \textbf{90.8}
      & 51.4
      & \textbf{10.8} \\

    LLLM $(908,256)$ re+vs
      & 43.6
      & 42.6
      & 87.7
      & \textbf{52.0}
      & 10.0 \\

    LLLM J3 $(945,256)$
      & \textbf{44.9}
      & \textbf{45.2}
      & 89.2
      & 51.5
      & 9.7 \\

    \midrule

    \multicolumn{6}{@{}l@{}}{%
      \colorbox{LLLMBlockGray}{%
        \parbox{\dimexpr\linewidth-2\fboxsep\relax}{%
          \rule{0pt}{2.5ex}%
          \textbf{600M-parameter models}%
        }%
      }%
    } \\

    Base-24L ($d_v=1620$)
      & 30.0
      & 31.7
      & 79.7
      & 42.9
      & 6.9 \\

    Base-40L ($d_v=1024$)
      & 33.5
      & 32.7
      & 83.3
      & 44.7
      & 7.0 \\

    LLLM $(1024,1024)$
      & \textbf{45.7}
      & \textbf{42.6}
      & \textbf{89.0}
      & \textbf{51.0}
      & \textbf{10.4} \\

    \bottomrule
  \end{tabularx}
  \vspace{2mm}
  \caption{
    \textbf{Finetuned mathematical-reasoning Acc.}
  }
  \label{tab:20b_finetuning_reasoning}
  \vspace{-5mm}
\end{wraptable}

We report accuracy on the GSM8K test set ($1{,}319$ problems) and, without further training, on
GSM-Hard ($1{,}319$), MultiArith ($600$), ASDiv ($2{,}305$), and a $1{,}000$-problem SVAMP subset (SVAMP-1K).
Table~\ref{tab:20b_finetuning_reasoning} evaluates how well the
20B-token-pretrained models transfer after finetuning on GSM8K-Aug.
At the 400M scale, every LLLM variant outperforms the matched Base-24L
Transformer across all five reasoning benchmarks.

LLLM$(945,256)$ J3 achieves the highest mean accuracy, improving from $36.02$ to $48.10$
($+12.08$ points), and obtains the strongest GSM8K and SVAMP-1K results.
The base-update model performs best on MultiArith and GSM-Hard, while
input injection and \texttt{vstream} produce the best ASDiv accuracy{;
these differences among Lattice variants are small (at most $3.2$ points)}. The trend persists at the 600M scale:
LLLM$(1024,1024)$ outperforms the 609M Base-40L Transformer on every
task and raises mean accuracy from $40.24$ to $47.74$ ($+7.50$ points),
despite using slightly fewer parameters. These results show that gains from horizontal latent capacity transfer
beyond pretraining perplexity to math reasoning.
Finetuned accuracy is not determined by pretraining quality alone: Base-40L and
LLLM$(908,256)$ with \texttt{input-injection}\,+\,\texttt{vstream} have the same FineWeb-Edu
perplexity ($14.40$) and nearly the same zero-shot average ($47.36$ vs.\ $47.38$), yet differ by
$10.1$ points on GSM8K ($33.5$ vs.\ $43.6$).
{Since each model is finetuned once, we do not interpret the small differences among
Lattice variants; the gap to the Transformers is larger and consistent, with} every Lattice model
ahead of every Transformer baseline by at least $10.1$ GSM8K points at both scales.

\definecolor{LLLMBlockGray}{RGB}{245,245,245}

\subsubsection{Recurrent Backbones}
\label{app:recurrent_experiments}

\begin{figure}[t]
    \centering
    \includegraphics[width=\linewidth]{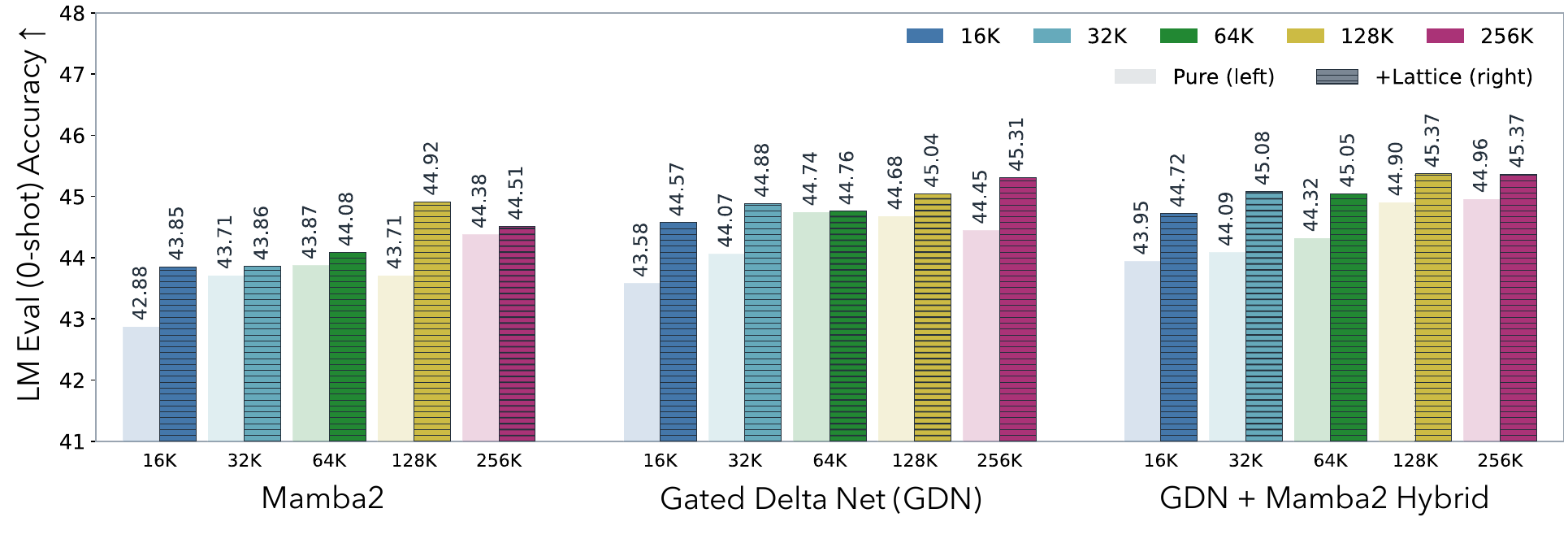}
    \vspace{-6mm}
    \caption{\textbf{Zero-shot downstream accuracy of recurrent backbones.}
    Nine-task average accuracy (\%) of pure and Lattice Mamba2, GDN, and no-SWA GDN--Mamba2
    hybrid models ($\sim$405M parameters, 10B FineWeb-Edu tokens) across native-state sizes.
    Lattice boosts pure backbone performance on the nine-task average results.
    See Sec.~\ref{sec:recurrent_backbone_experiments}.}
    \label{fig:recurrent_backbone_downstream_0shot}
    \vspace{-4mm}
\end{figure}

\begin{figure}[t]
    \centering
    \vspace{4mm}
    \includegraphics[width=\linewidth]{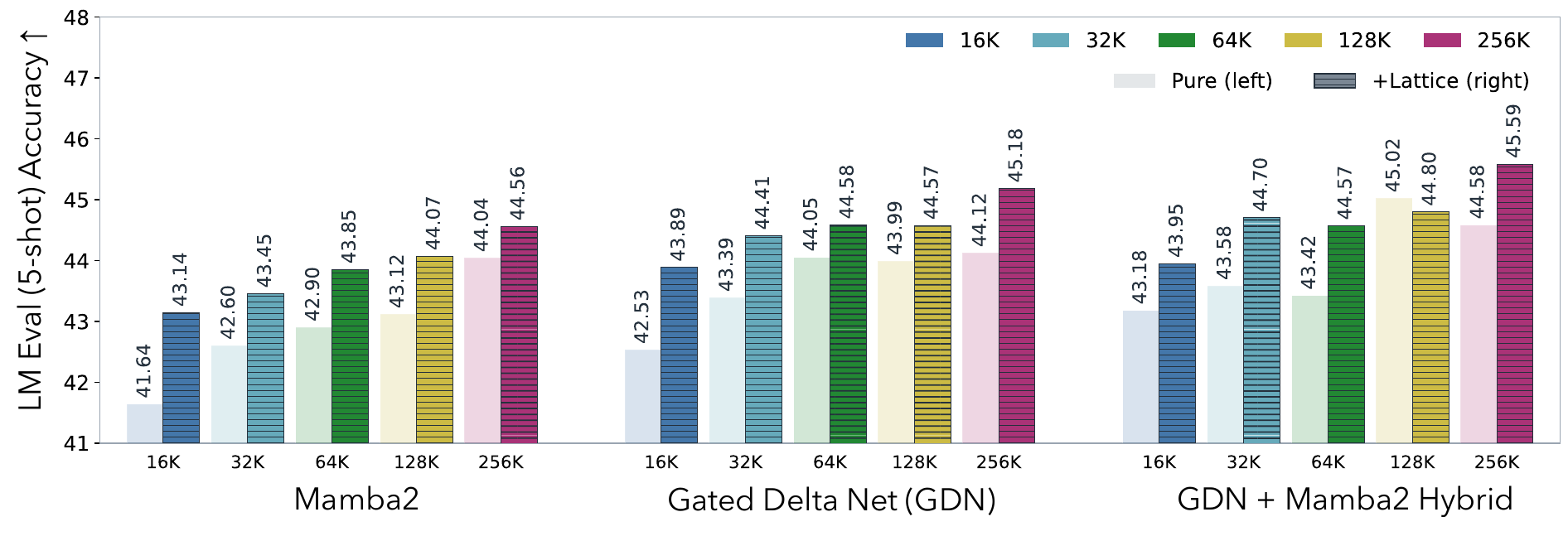}
    \vspace{-6mm}
    \caption{\textbf{Five-shot downstream accuracy of recurrent backbones.}
    Nine-task average accuracy (\%) of pure and Lattice Mamba2, GDN, and no-SWA GDN--Mamba2
    hybrid models ($\sim$405M parameters, 10B FineWeb-Edu tokens) across native-state sizes.
    Lattice boosts pure backbone performance on the nine-task average results.
    See Sec.~\ref{sec:recurrent_backbone_experiments}.}
    \label{fig:recurrent_backbone_downstream_5shot}
    \vspace{-2mm}
\end{figure}

\paragraph{Downstream evaluation.}
Figs.~\ref{fig:recurrent_backbone_downstream_0shot}
and~\ref{fig:recurrent_backbone_downstream_5shot} show the per-state-budget nine-task
averages summarized in Sec.~\ref{sec:recurrent_backbone_experiments}.
At each state budget and shot setting, the Lattice point is the best of
\texttt{input-injection}\,+\,\texttt{vstream} and two \texttt{vstream}-only variants that
reduce either $d_{\mathrm{model}}$ or the FFN width; all variants are within $0.1\%$ of the pure model's
parameter count.
Because this selection is post hoc, the gains should be read as those of the best coupling
rather than of a single preselected one.
The Lattice model is ahead in 29 of 30 pairs; the only regression is the hybrid at 128K
in the five-shot setting ($45.02 \rightarrow 44.80$).

\subsubsection{Residual Streams}
\label{app:residual_streams}
\begin{figure}[t]
    \centering
    \includegraphics[width=\linewidth]{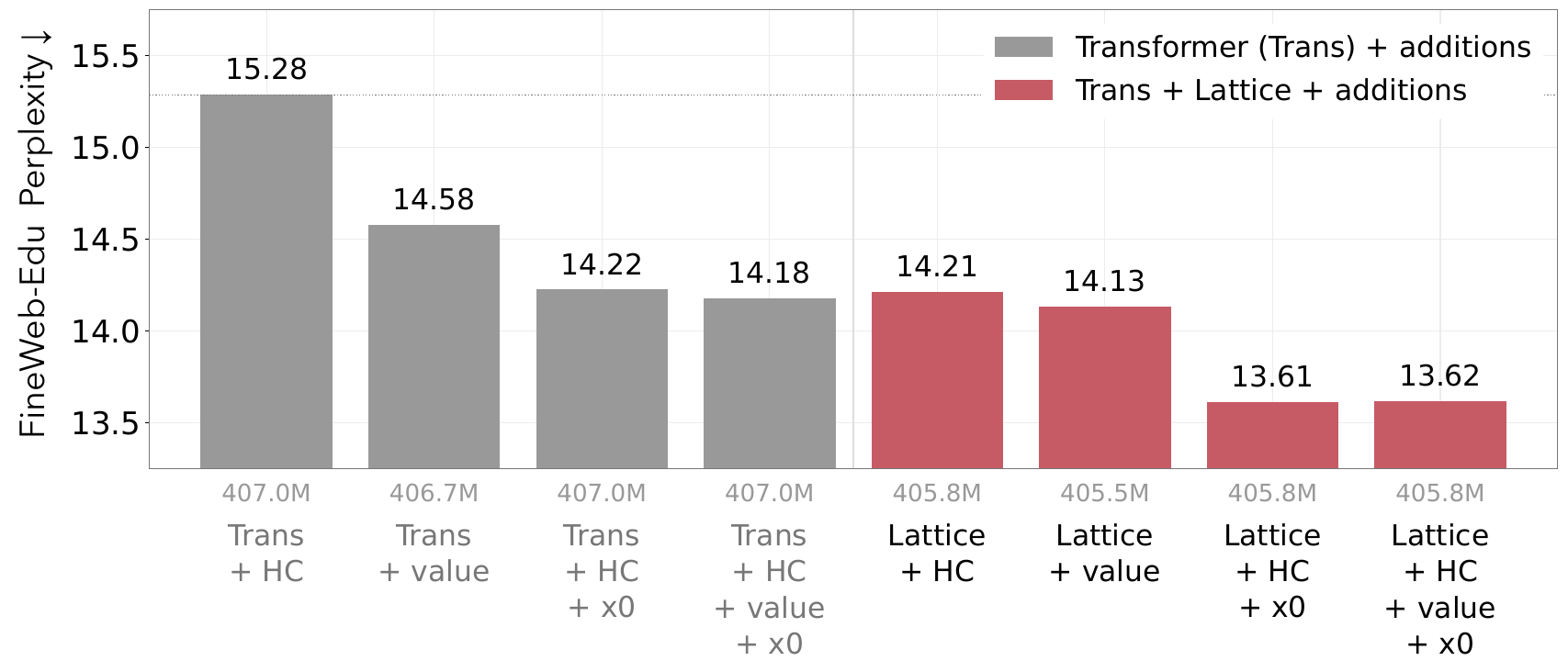}
    \vspace{-6mm}
    \caption{\textbf{Lattice of Latents vs.\ prior residual-stream enhancements: learned value residuals.} Complementing Fig.~\ref{fig:residual_stream_comparison}, we add learned value residuals (``value''), alone and on top of HC\,+\,x0, for Transformers (gray) and LLLMs (red). The Lattice lowers perplexity in every configuration ($3.0$--$7.0\%$).
    FineWeb-Edu validation perplexity, $\sim$400M models, trained on 20B tokens. See Appendix Sec.~\ref{app:training_residual} for details on the learned value residual.}
    \label{fig:residual_stream_comparison_v2}
    \vspace{-2mm}
\end{figure}

{\paragraph{Learned value residuals.}
The fixed value residual of Fig.~\ref{fig:residual_stream_comparison} mixes the first-layer and current values {with constant weights of $0.5$}; here we {learn these} weights (App.~\ref{app:training_residual}).
Fig.~\ref{fig:residual_stream_comparison_v2} reports validation
FineWeb-Edu perplexity after 20B tokens. Learning the weights improves on the fixed mix, e.g., Transformer\,+\,value $14.69\to14.58$, Lattice\,+\,value $14.26\to14.13$.
{Across all four configurations of Fig.~\ref{fig:residual_stream_comparison_v2}, the Lattice lowers perplexity} by $3.0$--$7.0\%$. 
On top of HC\,+\,x0, however, learned value residuals add little: $-0.3\%$ for the Transformer ($14.22\to14.18$) and none for the Lattice ($13.61\to13.62$). Lattice\,+\,HC\,+\,x0 therefore remains the best model overall.}

\subsection{Training Details}
\label{app:training}


Unless otherwise noted, all models use the same training recipe; only
model width and depth, token budget, and Lattice-specific settings vary
across runs.

{\subsubsection{Transformer Backbones}
\label{app:training_transformer}}

\paragraph{Backbone.}
We build on a LLaMA-style decoder with the GPT-NeoX tokenizer
(vocabulary size $V=50{,}304$). Layers use pre-RMSNorm
($\epsilon=10^{-5}$), rotary position embeddings (RoPE) with a context
length of $2048$, SwiGLU MLPs with intermediate size $2752$, and an
\emph{untied} output head. Attention uses $16$ query heads and $16$
key/value heads, each of dimension $64$, so the attention width
($16\times64=1024$) is fixed and decoupled from the residual-stream
(vertical-latent) width $d_{\mathrm{model}}$. We vary
$d_{\mathrm{model}}$$\in\{256,512,908,932,945,1024,\dots\}$ and the depth
$N\in\{22,24,40\}$ to sweep model size. Input embeddings are scaled by
$\sqrt{d_{\mathrm{model}}}$. Backbone attention and MLP projections have no biases, and all
weights not listed below use the standard $\mathcal{N}(0,0.02^2)$ initialization.

\paragraph{Optimizer.}
We use AdamW with $\beta_1{=}0.9$, $\beta_2{=}0.95$, weight decay $0.1$,
and gradient-norm clipping at $1.0$. Weight decay is not applied to biases, to the
gains of the backbone RMSNorm layers, or, following Mamba-2 practice, to the
horizontal-latent decay parameters $\mathrm{A\_log}$ and $\mathrm{dt\_bias}$; all other
parameters, including the learned initial state {$r_0$} and the normalization gains
inside the Lattice update, are decayed. All Lattice injection maps---the FiLM
$\gamma/\beta$ projections, input-injection and \texttt{vstream}
maps, Jacobi refiner's output projection, and output-head carry
projection---are zero-initialized. Consequently, the Lattice model is
functionally identical to a plain Transformer of the same width and depth at
initialization, and the horizontal latent acquires influence only through
training.

\paragraph{Learning-rate schedule.}
The peak learning rate is $1\times10^{-3}$ and is decayed with a cosine
schedule to a floor of $10\%$ of the peak
(\texttt{cosine\_with\_min\_lr},
$\texttt{min\_lr\_rate}{=}0.1$), following a linear warmup over the
first $2\%$ of training steps. We tuned the peak learning rate by grid search for both the Transformer and Lattice backbones.

\paragraph{Batching and schedule length.}
We use a global batch size of $512$ sequences of length $2048$
($\approx 1.05$M tokens per step), realized with a per-device batch size
of $8$ and gradient accumulation of $8$ across $8$ GPUs. Token budgets
correspond to $9{,}537$ steps for 10B tokens, $19{,}074$ for 20B,
and $28{,}611$ for 30B. 

\paragraph{Precision and kernels.}
Training uses \texttt{bfloat16} mixed precision, FlashAttention-2, and
gradient checkpointing; attention inside Lattice layers uses PyTorch's fused
scaled-dot-product attention. For numerical stability, the diagonal
gated-linear carry scan and its FiLM application are computed in
\texttt{fp32}.

\paragraph{Data.}
We pretrain on FineWeb-Edu, pre-tokenized with the NeoX
tokenizer and packed into contiguous $2048$-token sequences
(10B-, 20B-, and 30B-token subsets).
Perplexity is reported on a deduplicated, contamination-checked,
held-out FineWeb-Edu split, and out-of-domain perplexity on Pile, LAMBADA,
and Nemotron-CC (Sec.~\ref{sec:transformer_backbones}).

\paragraph{Lattice configuration.}
An LLLM is specified by its early/Lattice/late split $E/L/D$
(e.g., $E1/L20/D1$ or $E4/L16/D4$, with horizontal latents only in the
$L$ block), the horizontal-latent width
$d_r$ ($256$ for the 400M models; $d_r{=}d_{\mathrm{model}}$ for LLLM$(512,512)$ and LLLM$(1024,1024)$), groupwise FiLM in which each attention head
shares its multiplicative gain ({$\gamma$}) across $4$ channel groups, and the number of Jacobi sweeps
$J\in\{0,3\}$.

\begin{table}[h]
\centering
\small
\setlength{\tabcolsep}{6pt}
\renewcommand{\arraystretch}{1.05}
\begin{tabular}{@{}ll@{}}
\toprule
\textbf{Hyperparameter} & \textbf{Value} \\
\midrule
Optimizer
  & AdamW ($\beta_1{=}0.9,\ \beta_2{=}0.95$) \\
Weight decay
  & $0.1$ (none on biases/backbone norms/$\mathrm{A\_log}$/{$\mathrm{dt\_bias}$}) \\
Peak LR / floor
  & $1\times10^{-3}$ / $10\%$ of peak \\
LR schedule
  & cosine-with-min-lr, $2\%$ linear warmup \\
Grad clip
  & $1.0$ \\
Global batch / seq.\ length
  & $512$ / $2048$ ($\approx1.05$M tokens/step) \\
Per-device batch $\times$ grad-accum.
  & $8\times8$ on $8$ GPUs \\
{Steps (10B/20B/30B tokens)}
  & $9{,}537$ / $19{,}074$ / $28{,}611$ \\
Precision
  & bf16 (carry scan/FiLM in fp32), FlashAttention-2{/SDPA} \\
Vocab.\ / context
  & $50{,}304$ / $2048$ \\
Heads ($Q$/$KV$) $\times$ head dim.
  & $16/16\times64$ (attention width $1024$) \\
MLP intermediate
  & $2752$ (SwiGLU) \\
Embedding scale
  & {$\sqrt{d_{\mathrm{model}}}$} \\
\bottomrule
\end{tabular}
\caption{Training hyperparameters shared across all runs unless otherwise noted.}
\label{tab:training_hparams}
\end{table}

\subsubsection{Residual-Stream Enhancements}
\label{app:training_residual}

\paragraph{Setup.}
Fig.~\ref{fig:residual_stream_comparison} (Sec.~\ref{sec:residual_experiments}) compares ten
400M models trained on 20B FineWeb-Edu tokens ($19{,}074$ steps) with the Transformer recipe
above, changing only the residual stream.
The seven Transformer variants (plain, +HC, +mHC, +value, +x0, +HC\,+\,x0, +x0\,+\,value) use
the 24-layer, $d_{\mathrm{model}}{=}1024$ model. The three Lattice variants (+Lattice,
+Lattice\,+\,HC, +Lattice\,+\,HC\,+\,x0) use the \texttt{vstream}-only configuration
(E1/L20/D1, $d_{\mathrm{model}}{=}932$, $d_r{=}256$, no input injection), narrowed to stay
iso-parameter with the Transformer.
{Fig.~\ref{fig:residual_stream_comparison_v2} (App.~\ref{app:residual_streams}) adds four runs with learned value residuals, {two Transformers and their Lattice counterparts.}}

\paragraph{Input residual (x0).} Following nanochat~\citep{nanochat}, we compute
$h^0=\mathrm{RMSNorm}(\text{scaled token embedding})$ once, without a learned gain, and before
every layer $\ell$ (early, Lattice, and late) set $h\leftarrow\lambda_r^\ell h+\lambda_0^\ell h^0$.
The $2N$ scalars are initialized linearly in depth,
$\lambda_0^\ell=0.20-0.15\,\ell/(N{-}1)$ and $\lambda_r^\ell=1.15-0.10\,\ell/(N{-}1)$,
and trained in separate AdamW groups: $\lambda_0$ at $25\times$ the base learning rate
($2.5\times10^{-2}$) without weight decay, and $\lambda_r$ at $0.25\times$
($2.5\times10^{-4}$) with weight decay $0.05$; both follow the same warmup and cosine schedule.

\paragraph{Value residual.} In Fig.~\ref{fig:residual_stream_comparison}, we use the fixed (constant) ResFormer variant~\citep{zhou2025valueresiduallearning}: every layer
after the first replaces its values with $\tfrac12 v^1+\tfrac12 v^\ell$, with no added
parameters.
{The learned variant (all results in Fig.~\ref{fig:residual_stream_comparison_v2}) instead uses $\alpha^\ell v^1+\beta^\ell v^\ell$, with two unconstrained
scalars per layer initialized to $0.5$, adding $2(N{-}1)$ parameters ($46$ for the Transformer,
$42$ for the Lattice). Like the x0 scalars, $\alpha^\ell$ is trained at $25\times$ the base learning
rate without weight decay, and $\beta^\ell$ at $0.25\times$ with decay $0.05$.}

\paragraph{Hyper-connections.} HC~\citep{zhu2025hyperconnections} and mHC~\citep{xie2026mhcmanifoldconstrainedhyperconnections} use $n{=}4$
residual streams, each initialized with a copy of the token embedding and summed before the
final RMSNorm.
{Each attention and MLP sub-layer has its own router with three sets of weights:
\emph{read} weights that mix the $n$ streams into the sub-layer input, \emph{write} weights
that distribute the sub-layer output back to the streams, and an $n\times n$ \emph{mixing}
matrix that recombines the streams.
In HC, following the dynamic formulation of \citet{zhu2025hyperconnections} (App.~J), each
weight is a static term plus a token-dependent term computed from the LayerNorm-normalized
streams. The static terms are initialized so that each sub-layer reads from one stream and
the mixing matrix is the identity; the token-dependent terms are zero-initialized and scaled
by learnable gates initialized to $0.01$. The attention and MLP output projections are
scaled by $1/\sqrt{n}=0.5$ at initialization.
In mHC, following \citet{xie2026mhcmanifoldconstrainedhyperconnections} (Eqs.~7--9), a single
linear map of the RMS-normalized, flattened streams produces all three: read weights
$\sigma(\cdot)$, write weights $2\sigma(\cdot)$, and mixing logits, which $20$
Sinkhorn--Knopp iterations project toward a doubly stochastic matrix. The static biases are
zero for the read and write weights and the identity for the mixing logits.
In both, the static router weights are excluded from weight decay.}

\paragraph{Composition with the Lattice.} We combine the Lattice with HC: each Lattice layer
keeps a single horizontal latent, shared by all four streams, which reads the
attention router's mixed input and steers the attention and MLP exactly as in the
single-stream model.
{The horizontal latent does not modulate the HC read, write, or mixing weights;
we leave this to future work.}
Parameter counts are $406.65$\,M (Transformer), $407.03$\,M (HC),
$411.36$\,M (mHC), $405.52$\,M (Lattice), and $405.85$\,M (Lattice\,+\,HC); x0 adds only
$2N$ scalars{,} the fixed value residual {none, and the learned value residual $2(N{-}1)$}.

\begin{table}[t]
\centering
\scriptsize
\setlength{\tabcolsep}{4pt}
\renewcommand{\arraystretch}{1.05}
\begin{tabular}{@{}lccccc@{}}
\toprule
\textbf{Native state} & \textbf{16K} & \textbf{32K} & \textbf{64K} & \textbf{128K} & \textbf{256K} \\
\midrule
GDN $(H,d_k,d_v)$ & $16,16,64$ & $16,32,64$ & $16,64,64$ & $8,128,128$ & $4,256,256$ \\
\quad $d_{\mathrm{model}}$ pure / Lattice & $1120/998$ & $1086/963$ & $1024/898$ & $1024/899$ & $1024/900$ \\
Mamba2 $(H,d_{\mathrm{state}},d_{\mathrm{head}})$ & $16,16,64$ & $16,32,64$ & $16,64,64$ & $16,128,64$ & $4,256,256$ \\
\quad $d_{\mathrm{model}}$ pure / Lattice & $1155/1036$ & $1153/1034$ & $1148/1029$ & $1139/1019$ & $1122/1003$ \\
Hybrid $d_{\mathrm{model}}$ pure / Lattice & $1137/1028$ & $1118/1008$ & $1082/971$ & $1079/969$ & $1071/961$ \\
\bottomrule
\end{tabular}
\caption{\textbf{Recurrent-backbone geometries and widths.} Native state per mixer and
residual width of the pure and \texttt{input-injection}\,+\,\texttt{vstream} Lattice
models (all $\sim$405M parameters). Hybrid GDN and Mamba2 mixers use the GDN and Mamba2
geometries of the same column. Triadic GDN pairs use joint geometries
$(H,Ed_k,d_v)=(16,16,64)$, $(16,32,64)$, $(16,64,64)$, $(32,64,32)$, $(16,128,64)$, and
$(8,256,128)$.}
\label{tab:recurrent_widths}
\end{table}

\subsubsection{Recurrent Backbones}
\label{app:training_recurrent}

\paragraph{Common setup.}
All recurrent models (Sec.~\ref{sec:recurrent_backbone_experiments}) are attention-free
stacks of $22$ pre-RMSNorm layers, each a token mixer followed by a SwiGLU MLP
(intermediate size $2752$), with the GPT-NeoX vocabulary ($V=50{,}304$), untied embeddings, and
no positional encoding. All mixers run on the Triton kernels of the flash-linear-attention
library (\texttt{fla}, v0.4.1)\footnote{\texttt{github.com/fla-org/flash-linear-attention}; we add
only a small PyTorch-2.4 compatibility shim and modify no kernel or layer code.}. Baseline
and Lattice models share the same mixer code; the Lattice versions only add the steering
described below. Native-state geometries and widths are listed in Tab.~\ref{tab:recurrent_widths}.

\paragraph{GDN.}
We use \texttt{fla}'s \texttt{GatedDeltaNet} layer~\citep{yang2025gateddeltanetworksimproving}
unchanged: short causal convolutions on $q,k,v$, in-kernel $L_2$ normalization of $q$ and $k$,
per-head decay $\alpha$ and write strength $\beta$, a gated output RMSNorm, and the chunked
training kernel. The state per head is $d_k\times d_v$, varied through the number of heads $H$
and the per-head key and value dimensions.

\paragraph{Mamba2.}
We implement the Mamba-2 SSD layer~\citep{dao2024transformersssmsgeneralizedmodels} on
\texttt{fla}'s \texttt{chunk\_simple\_gla} kernel rather than the \texttt{mamba\_ssm} package:
a joint input projection to $[z,\,xBC,\,\Delta]$, a size-4 depthwise causal convolution with
SiLU on $xBC$, a scalar per-head decay $\Delta A$ with $A=-\exp(A_{\log})$ and
$A_{\log}$ initialized to $\log(1,\dots,H)$, values $x\Delta$, a $D$ skip connection, a $z$-gated
RMSNorm, and the standard log-uniform $\Delta$-bias initialization on $[10^{-3},10^{-1}]$.
The inner width is $H\cdot d_{\mathrm{head}}=1024$ with one $B/C$ group, and the state per head is
$d_{\mathrm{state}}\times d_{\mathrm{head}}$.

\paragraph{Triadic GDN.}
The official triadic linear attention kernels support only a per-head key dimension of $128$
on Hopper/Blackwell GPUs and provide no fallback, so we reimplement the layer inside our
GDN stack. Each head writes with the joint key $\kappa=k_2\otimes k$ and reads with the joint
query $\pi=q_2\otimes q$, where $k,q\in\mathbb{R}^{d_k}$ are $L_2$-normalized GDN-style factors
and $k_2,q_2\in\mathbb{R}^{E}$ ($E{=}2$ slices) are $L_2$-normalized softplus projections of the
input. The recurrence is the standard gated delta rule on the joint key dimension $Ed_k$,
computed with \texttt{fla}'s \texttt{chunk\_gated\_delta\_rule}. Our one simplification relative
to the original is a scalar per-head decay shared across slices.

\paragraph{GDN--Mamba2 hybrid.}
We follow the Gated DeltaNet-H2 layout~\citep{yang2025gateddeltanetworksimproving} with
only GDN and Mamba layers: $11$ blocks of [Mamba2, MLP, GDN, MLP], i.e., $22$ mixers.
Its Mamba2 and GDN layers are the same as in the pure backbones above.

\paragraph{Adding the Lattice.}
Each Lattice layer adds the same horizontal-latent update as in the Transformer, with
$d_r=256$, input injection, and \texttt{vstream}. The pure GDN, Mamba2, and Triadic GDN
backbones use E1/L20/D1, so $20$ of the $22$ mixer layers carry a latent. The hybrid applies
the same split at the level of its $11$ blocks (E1/L9/D1): each of the $9$ Lattice blocks
gives its Mamba2 mixer and its GDN mixer their own latent, for $18$ latents in total.
Instead of attention, the latent steers the mixer by FiLM,
$x\leftarrow x\odot(1+\tanh\gamma)+\beta$, applied to every quantity that controls
native-memory writes and reads: for GDN, $q,k,v$ (after the short convolution and before
$L_2$ normalization), the decay and write-strength logits (before softplus and sigmoid),
and the output-gate input; for Mamba2, $C$, $B$, $x$ (the analogues of $q,k,v$), the
per-head $\Delta$ logit (before softplus), and $z$; for Triadic GDN, additionally the
$k_2,q_2$ logits before softplus. The MLPs, the output head, and the native-state geometry
are unchanged, and all FiLM maps are zero-initialized, so each Lattice model equals its pure
counterpart of the same width at initialization. To keep each pair iso-parameter, we
reduce $d_{\mathrm{model}}$ of the Lattice model (Tab.~\ref{tab:recurrent_widths}). The two
\texttt{vstream}-only variants used for downstream selection have no input injection and
instead either reduce $d_{\mathrm{model}}$ at FFN width $2752$ or keep
$d_{\mathrm{model}}{=}1024$ and reduce the FFN width.

\paragraph{Training.}
All recurrent runs use a standalone PyTorch DDP trainer on $8$ GPUs with the Transformer
recipe above (AdamW, peak learning rate $10^{-3}$, cosine decay to $10\%$, weight decay $0.1$,
clipping $1.0$, global batch $512\times2048$, $9{,}537$ steps $\approx$ 10B FineWeb-Edu
tokens, \texttt{bfloat16}, gradient checkpointing, seed $42$), with two differences:
linear warmup over the first $10\%$ of steps, and weight decay applied only to matrices
other than the token embedding (no decay on embeddings, normalization gains, biases,
$r_0$, $A_{\log}$, $\Delta$/dt biases, or $D$). Pure and Lattice models within each pair
share this recipe exactly.

\end{document}